\documentclass{article}

\usepackage[preprint]{neurips_2025}

\usepackage[utf8]{inputenc} % allow utf-8 input
\usepackage[T1]{fontenc}    % use 8-bit T1 fonts
\usepackage{hyperref}       % hyperlinks
\usepackage{url}            % simple URL typesetting
\usepackage{booktabs}       % professional-quality tables
\usepackage{amsfonts}       % blackboard math symbols
\usepackage{nicefrac}       % compact symbols for 1/2, etc.
\usepackage{microtype}      % microtypography
\usepackage{xcolor}         % colors

\usepackage{multirow}
\usepackage{colortbl}
\usepackage{longtable}

\usepackage{bbm}
\usepackage{bbding}
\usepackage{amsmath}
\usepackage{graphicx}
\usepackage{subcaption}
\usepackage{tikz}
\usepackage{pgfplots}
\usepackage{pgfplotstable}
\usepackage{tcolorbox}
\usepackage{makecell}
\usepackage{amssymb}
\usepackage{pifont}
\definecolor{color1}{RGB}{200,230,240}
\definecolor{color2}{RGB}{235,245,255}
\definecolor{lavender}{RGB}{220,220,250}

\usepackage{booktabs}

\usepackage{multirow}
\usepackage{lineno}
\usepackage{graphicx}
\usepackage{wrapfig}
\usepackage{graphicx}
\usepackage{tabularx}
\usepackage{array}  % 导入 array 宏包

\title{Benchmarking Multimodal Knowledge Conflict for Large Multimodal Models}

\author{%
  Yifan Jia$^{1}$\thanks{Equal contribution}, Kailin Jiang$^{2}$*, Yuyang Liang$^{1}$, Qihan Ren$^{3}$,
  Yi Xin$^{4}$, \\ \textbf{Rui Yang$^{1}$,  Fenze Feng$^{1}$,   Mingcai Chen$^{5}$,  Hengyang Lu$^{6}$, Haozhe Wang$^{7}$,} \\ \textbf{Xiaoye Qu$^{8}$,   Dongrui Liu$^{8}$, Lizhen Cui$^{1}$, Yuntao Du$^{1}$\thanks{Corresponding author}}  \\
\small $^1$ Joint SDU-NTU Centre for Artificial Intelligence Research\&School of Software , Shandong University \\
\small $^2$ University of Science and Technology of China 
\small $^3$ Shanghai Jiaotong University 
\small $^4$ Nanjing University,  \\
\small $^5$ Nanjing University of Posts and Telecommunications
\small $^6$ Jiangnan University \\
\small $^7$ The Hong Kong University of Science and Technology
\small $^8$ Shanghai AI Laboratory \\
}

\begin{document}

\maketitle

\begin{abstract}

Large Multimodal Models (LMMs) face notable challenges when encountering multimodal knowledge conflicts, particularly under retrieval-augmented generation (RAG) frameworks, where the contextual information from external sources may contradict the model’s internal parametric knowledge, leading to unreliable outputs. However, existing benchmarks fail to reflect such realistic conflict scenarios. Most focus solely on intra-memory conflicts, while context-memory and inter-context conflicts remain largely investigated. Furthermore, commonly used factual knowledge-based evaluations are often overlooked, and existing datasets lack a thorough investigation into conflict detection capabilities.
To bridge this gap, we propose \textbf{MMKC-Bench}, a benchmark designed to evaluate factual knowledge conflicts in both context-memory and inter-context scenarios. MMKC-Bench encompasses three types of multimodal knowledge conflicts and includes 1,573 knowledge instances and 3,381 images across 23 broad types, collected through automated pipelines with human verification. We evaluate three representative series of LMMs on both model behavior analysis and conflict detection tasks. Our findings show that while current LMMs are capable of recognizing knowledge conflicts, they tend to favor internal parametric knowledge over external evidence. We hope MMKC-Bench will foster further research in multimodal knowledge conflict and enhance the development of multimodal RAG systems.
The source code is available at \url{https://github.com/MLLMKCBENCH/MLLMKC}.

\end{abstract}

\vspace{-3mm}

% \vspace{-3mm}
\section{Introduction}

% The rapid advent of large multimodal models (LMMs) and large language models (LLMs) has demonstrated excellent performance in various multimodal understanding, generation, and reasoning tasks~\cite{bai2025qwen2,chen2024internvl,liu2023visual,cui2024survey}. However, static LMMs or LLMs often suffer from incorrect or outdated information and hallucinations. To overcome this challenge, some techniques, like retrieval-augmentation generation (RAG)~\cite{fan2024survey,mei2025survey} have been proposed to utilize up-to-date knowledge from external sources for LMMs. Although effective, the RAG framework often encounters a ``knowledge conflict'', where external sources' contexts may conflict with the parametric knowledge of the model~\cite{xu2024knowledge}, and recent work has shown that knowledge conflict could significantly affect the trustworthiness and reliability of models~\cite{wang2023resolving}, thus it is necessary to understand the analysis of the model behavior under conflict scenarios. To achieve this, several datasets have been proposed to analyze the behavior of knowledge conflict in both textual~\cite{hou2024wikicontradict,su2024texttt,wang2023resolving} and multimodal domains~\cite{liu2024insight,shao2024cognition,zhu2024unraveling}.

The rapid advancement of large multimodal models (LMMs) and large language models (LLMs) has led to remarkable performance across a wide range of multimodal understanding, generation, and reasoning tasks~\cite{bai2025qwen2, chen2024internvl, liu2023visual, cui2024survey,su2025openthinkimg}. Despite their impressive capabilities, static LMMs and LLMs often suffer from limitations such as outdated or incorrect knowledge and hallucinations. To address these issues, retrieval-augmented generation (RAG) techniques have been introduced~\cite{fan2024survey, mei2025survey}, which enhance model outputs by incorporating up-to-date information from external sources. However, this paradigm introduces the challenge of knowledge conflict, where the retrieved contextual knowledge may contradict the model’s internal (parametric) knowledge~\cite{xu2024knowledge}. Recent studies have demonstrated that such conflicts can undermine the trustworthiness and reliability of model predictions~\cite{wang2023resolving}, highlighting the need for a deeper understanding of model behavior under conflicting knowledge scenarios. To facilitate this, several benchmark datasets have been developed to study knowledge conflict both in textual contexts~\cite{hou2024wikicontradict, su2024texttt, wang2023resolving} and in multimodal domains~\cite{liu2024insight, shao2024cognition, zhu2024unraveling}.

 % previous work~\cite{hou2024wikicontradict,su2024texttt} has mainly explored knowledge conflict for large language models, while the impact of knowledge conflicts for large multimodal models based on multimodal input has been rarely explored. 
% under-explored

% As shown in the survey~\cite{}, there are usually three kinds of conflict scenarios, namely
%  intra-memory conflicts~\cite{xu2024knowledge},
%  context-memory conflicts, and inter-context conflicts.  The intra-memory conflict refers to the inconsistencies between the LLM’s parametric knowledge present in the pretraining data, which is caused by model pre-training.  While the last two conflicts often occur in inference time.  The context-memory conflicts refer to the retrieved context knowledge (including user prompts, dialogue history, and retrieved documents) conflict with the parametric knowledge (memory), inter-context conflicts refer to the contradictions among the retrieved contextual knowledge.

As outlined in the survey~\cite{xu2024knowledge}, knowledge conflicts in LLMs typically fall into three categories: intra-memory conflicts, context-memory conflicts, and inter-context conflicts. Intra-memory conflict arises from inconsistencies within the model’s own parametric knowledge, which are often introduced during pretraining due to contradictory or noisy pretraining data. In contrast, context-memory and inter-context conflicts typically emerge during inference. Context-memory conflict occurs when external contextual information, such as user prompts or retrieved documents, contradicts the model’s internal (parametric) knowledge. Inter-context conflict refers to inconsistencies among the external contextual sources themselves. These latter forms of conflict are particularly relevant in RAG settings and present unique challenges for ensuring the reliability and consistency of model outputs.

% Existing multimodal knowledge conflict datasets focus on various aspects, such as commonsense-based knowledge conflict under context-memory scenario ~\cite{liu2024insight}, cognition and perception knowledge conflicts under intra-memory scenario by ulitzing different reality to deal with the same question(e.g. OCR vs VQA) ~\cite{shao2024cognition}, and cross-modality knowledge conflict intra-memory scenario by representing the same question with both textual and multimodal format ~\cite{zhu2024unraveling}. While showing insightful results, these works has not concentrated on realistic conflict scenarios under the RAG framework. First,  the existing dataset has mainly focused on intra-memory conflicts, while context-memory conflicts and inter-context conflicts have been rarely explored.   Second,
%  in realistic scenarios, we often require factual knowledge from an external source, such as what is the object in the image or the age of the person in the image, which is ignored by existing datasets. Lastly, existing datasets in multimodal scenarios is only designed to observe the model behavior and lack investigation that whether the model could effectively acknowledge the conflict.

Several multimodal knowledge conflict datasets have been proposed to investigate a variety of perspectives, such as commonsense-based knowledge conflicts under the context-memory scenario~\cite{liu2024insight}, cognition and perception conflicts under the intra-memory scenario by leveraging different model capability to answer the same question (e.g., OCR vs. VQA)\cite{shao2024cognition}, and cross-modality conflicts in intra-memory settings by framing the same question in both textual and multimodal formats\cite{zhu2024unraveling}. While these efforts provide valuable insights, they fall short of capturing realistic conflict scenarios within the RAG framework. First, existing datasets mainly focus on intra-memory conflicts, with limited attention paid to context-memory and inter-context conflicts. Second, in practical scenarios, external factual knowledge, such as identifying objects in an image or determining a person’s age, is often necessary but remains underrepresented in current datasets. Lastly, current multimodal conflict benchmarks are primarily designed to observe model behavior, rather than evaluating whether models can effectively detect and acknowledge the presence of knowledge conflict.

% To investigate the impact on large language models, some benchmarks have been proposed~\cite{}. To construct conflict datasets for controlled experiments, previous works primarily utilize word-level substitution~\cite{}, language model generation methods~\cite{}, or their combination~\cite{} to create conflicts~\cite{}.  

\begin{figure}[t!]
  \vspace{-3mm}
  \centering
\includegraphics[width=0.9\linewidth]{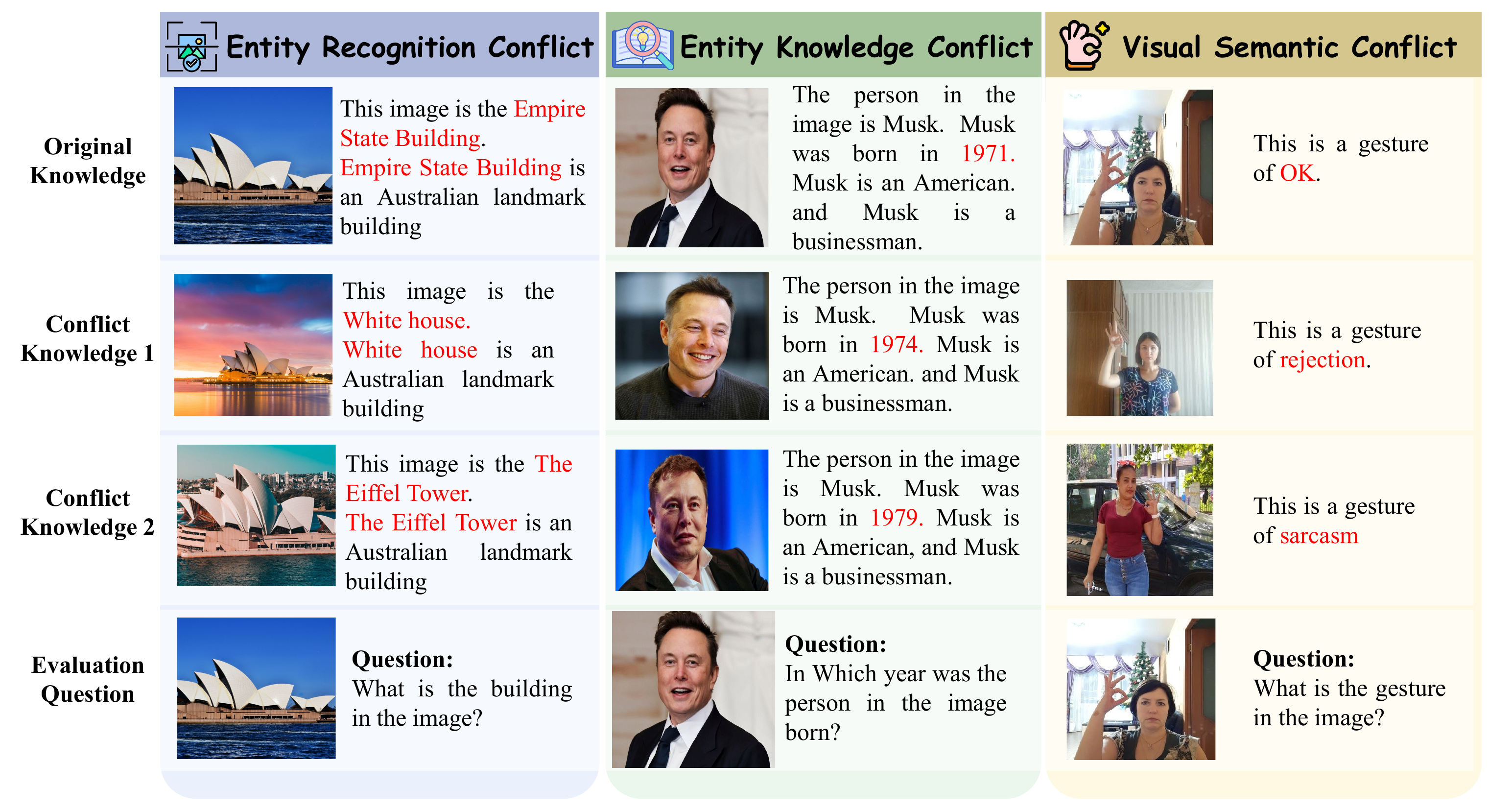}
  \caption{Three types of multimodal knowledge conflict in MMKC-Bench. It is noted that the original knowledge is shown to help understand what the conflict is,  and is not contained in the dataset.}
  \label{fig:motivation}
  \vspace{-6mm}
\end{figure}

To address these gaps, we propose MMKC-Bench, \textbf{a multimodal knowledge conflict benchmark aimed at evaluating factual knowledge conflicts under both context-memory and inter-context scenarios}. As illustrated in Fig.~\ref{fig:motivation}, MMKC-Bench focuses on three representative types of multimodal conflicts: entity recognition conflict, entity knowledge conflict, and visual semantic conflict. The entity recognition and visual semantic conflicts target inconsistencies in entity identification and complex visual understanding, while the entity knowledge conflicts emphasize factual inconsistencies related to specific attributes or quantitative data. In addition to analyzing model behavior in the presence of conflict, MMKC-Bench also investigates whether models can detect and perceive conflicts at both coarse-grained (given whole evidence) and fine-grained (given a subset of the evidence) levels.

To construct the benchmark, we first curate original multimodal knowledge from a variety of sources, including Wikipedia, Google Images, and existing datasets. We then employ large language models (LLMs) to generate conflicting knowledge through counterfactual editing, which involves modifying the entity name, semantic content, or entity-related factual information. Based on the constructed knowledge pairs, we use LLMs to generate both multiple-choice and open-ended evaluation questions, along with candidate answers. All generated questions, answers, and the preceding data construction steps undergo rigorous human verification, with samples being filtered or revised as needed to ensure quality and accuracy. The final MMKC-Bench dataset comprises 1,573 knowledge instances and 3,381 images, spanning 23 broad knowledge categories.

We conduct experiments on nine representative LMMs from three prominent model families known for their strong performance in multi-image reasoning: Qwen2.5-VL~\cite{bai2025qwen2}, InternVL3~\cite{zhu2025internvl3}, and GPT-4o mini~\cite{hurst2024gpt}. Our evaluation covers both model behavior analysis and conflict detection tasks. The experimental findings reveal several key insights:
(1) LMMs tend to rely more heavily on internal (parametric) knowledge than on external evidence, a behavior that contrasts with previously observed trends in LLMs~\cite{su2024texttt,hou2024wikicontradict};
(2) LMMs are more sensitive to knowledge-level conflicts (e.g., entity knowledge) than to recognition-level conflicts (e.g., entity identification);
(3) Larger models  show a stronger promoting effect across all conflict types;
(4) LMMs are capable of accurately identifying the presence of conflict in both coarse-grained and fine-grained scenarios.

To sum up, the contribution of this work is summarized as follows,
\begin{itemize}
  \item  We propose MMKC-Bench, a multimodal knowledge conflict benchmark focusing on factual knowledge conflict under both context-memory and inter-context scenarios.
  \item   We propose a novel pipeline to construct the benchmark that collects original knowledge, generates conflict knowledge and produce evaluation with two question formats. 
  \item   Extensive experiments with various models under both context-memory and inter-context for behavior understanding and conflict detection are conducted, revealing several characteristics of existing LMMs.
\end{itemize}

% \newpage
\vspace{-3mm}
\section{Related work}
\vspace{-2mm}
\subsection{Large Multimodal Model}
\vspace{-1mm}

The development of LMMs has significantly advanced the integration of visual and textual information, enabling more sophisticated cross-modal understanding and reasoning. Modern LMMs are typically composed of three core components: a language encoder, a vision encoder, and cross-modality alignment modules~\cite{caffagni2024revolution}. The language encoder is usually based on large language models such as LLaMA~\cite{grattafiori2024llama,touvron2023llama} and Qwen~\cite{yang2024qwen2,yang2024qwen2}, while the vision encoder often adopts architectures like ViT~\cite{dosovitskiy2020image}. Cross-modality alignment modules play a crucial role in integrating visual features into textual representations, allowing the language encoder to effectively interpret visual signals.
Based on this architecture, many state-of-the-art LMMs have been developed, including Qwen2.5-VL~\cite{bai2025qwen2}, InternVL2.5~\cite{chen2024expanding}, LLaVA~\cite{liu2023visual}, and LLaVA-OneVision~\cite{li2024llava}. In parallel, a range of training strategies have been introduced to strengthen cross-modal alignment. For instance, Qwen-VL integrates a visual receptor and employs a three-stage training pipeline to boost performance. InternVL adopts a native multimodal pre-training paradigm, jointly learning visual and linguistic knowledge from diverse sources. Meanwhile, LLaVA and LLaVA-Next enhance vision-language integration with improved visual grounding and reasoning capabilities.
These advancements have led to impressive performance across a wide spectrum of multimodal tasks~\cite{kil2024mllm,huang2024survey}, highlighting the rapid progress and evolving design of LMM architectures and training methodologies. Besides, some LLM Agents~\cite{gao2024clova,fan2024videoagent,gupta2023visual} utilizes visual tools to solve multimodal tasks, which is beyond our scope.

% \vspace{-2mm}
\subsection{Knowledge Conflict in LLMs}
% \vspace{-1mm}

% in LLM

% Understadning

% 缓解

% in VLM 

% Knowledge conflict refers to the discrepancies between the contexts and the model’s parametric knowledge~\cite{chen2022rich,xie2023adaptive}.  The cause of knowledge is from temporal misalignment, misinformation, bias in Corpora, and so on. Knowledge conflicts in LLMs are broadly categorized into context-memory conflicts, inter-context conflicts, and intra-memory conflicts. 
% While retrieved conflicts dominate research due to the rise of RAG, embedded conflicts, such as outdated or conflicting memorized information, receive less attention.   Most datasets rely on synthetic generation, either through entity substitution~\cite{longpre2021entity,wang2023resolving} or LLM-based contradiction creation~\cite{ying2023intuitive}. And Conflictbank~\cite{su2024texttt} constructs the largest evaluation datasets up to now. These benchmarks simplify conflicts into binary factual mismatches, lacking real-world complexity. Recent efforts like WikiConflict~\cite{hou2024wikicontradict} extract contradictions from Wikipedia, offering more natural examples. Besides, the following work also focuses on conflict detection\cite{wang2023resolving,li2023contradoc} and conflict mitigation\cite{gekhman2023trueteacher,chuang2023dola,cheung2023factllama} for better utilization.

Knowledge conflict refers to discrepancies between contextual inputs and a model’s internal parametric knowledge~\cite{chen2022rich,xie2023adaptive}. Such conflicts often arise from temporal misalignment or misinformation in training corpora. In the context of LLMs, knowledge conflicts are typically categorized into three types: context-memory conflicts, inter-context conflicts, and intra-memory conflicts.
With the rise of retrieval-augmented generation (RAG), research has largely focused on context-memory conflicts, while other conflicts have received comparatively less attention. Most existing datasets are synthetically constructed, either via entity substitution~\cite{longpre2021entity,wang2023resolving} or LLM-generated contradictions~\cite{ying2023intuitive}. Among them, ConflictBank~\cite{su2024texttt} is currently the largest benchmark, though its design simplifies conflicts into binary factual mismatches, limiting its reflection of real-world complexity. More recent datasets, such as WikiConflict~\cite{hou2024wikicontradict}, attempt to address this limitation by extracting natural contradictions from Wikipedia articles.
Beyond dataset construction, several studies have also explored conflict detection~\cite{wang2023resolving,li2023contradoc} and conflict mitigation~\cite{gekhman2023trueteacher,chuang2023dola,cheung2023factllama}, improving the reliability of LLM.

% \vspace{-2mm}
\subsection{Knowledge Conflict in LMMs}
% \vspace{-1mm}

% Some work also explores knowledge conflicts in LMMs from various aspects. Existing benchmarks include commonsense-based knowledge conflicts under the context-memory scenario~\cite{liu2024insight}, cognition and perception conflicts under the intra-memory scenario by utilizing different model capability to answer the same question (e.g., OCR vs. VQA)\cite{shao2024cognition}, and cross-modality conflicts in intra-memory settings by framing the same question in both textual and multimodal formats\cite{zhu2024unraveling}. 
% While these efforts provide valuable insights, they fall short of capturing realistic conflict scenarios within the RAG framework.  However, existing datasets mainly focus on intra-memory conflicts,  limited attention is given to context-memory and inter-context conflicts. Second, in practical scenarios, external factual knowledge, such as identifying objects in an image or determining a person’s age, is often necessary and widely used but remains underrepresented in current datasets. Lastly, current multimodal conflict benchmarks are primarily designed to observe model behavior, lacking evaluating whether models can effectively detect and acknowledge the presence of knowledge conflict.

Several studies have also explored knowledge conflicts in LMMs from different perspectives. For example, existing benchmarks have examined commonsense-based knowledge conflicts in context-memory scenarios~\cite{liu2024insight}, cognition-perception conflicts in intra-memory settings by leveraging different model capabilities to answer the same question (e.g., OCR vs. VQA)~\cite{shao2024cognition}, and cross-modality conflicts within intra-memory scenarios by comparing model responses to the same question framed in both textual and multimodal formats~\cite{zhu2024unraveling}.
While these efforts offer valuable insights, they fall short in capturing realistic conflict scenarios within the RAG framework. Specifically, current benchmarks predominantly focus on intra-memory conflicts, with limited attention to context-memory and inter-context conflicts. Moreover, in practical applications, resolving external factual knowledge conflicts, such as identifying objects in images or estimating a person’s age, is often crucial, yet remains underrepresented in existing datasets. Finally, most current multimodal conflict benchmarks are primarily designed to observe model behavior, lacking an explicit focus on evaluating whether models can detect and acknowledge the presence of knowledge conflicts.

% \vspace{-3mm}
\section{Problem Definition}
\subsection{Original and Conflict Knowledge Representation}

MMKC-Bench focuses on factual knowledge based knowledge conflict and encompasses three types:  entity recognition conflict, entity knowledge conflict, and visual semantic conflict. For these types, each original piece of original knowledge is represented in a unified format $k=(i,d)$, where $i$ denotes an image of the entity or semantic action, and $d$ is the corresponding textual description. To construct conflicting instances, the original knowledge is modified to form $k_c$.
 It takes the form $k_c = (i_c, d_c)$, where $i_c$ is another image of the same entity or action, and $d_c$ is a conflicting description. The examples of each type are shown in Fig~\ref{fig:motivation}.

\subsection{Multimodal Knowledge Conflict Types}

To comprehensively reflect real-world scenarios, MMKC-Bench includes three types of multimodal knowledge conflicts: entity recognition conflict, entity knowledge conflict, and  visual semantic conflict.

\textbf{Entity Recognition Conflict} simulates cognitive inconsistencies where different sources identify the same entity differently. This is achieved by keeping the entity image unchanged while replacing the entity name in the description with that of another entity of the same type. For example, as shown in Fig~\ref{fig:motivation}, describing the Empire State Building image using ``Eiffel Tower” or ``White House” as the entity name.

\textbf{Visual Semantic Conflict} addresses inconsistencies in interpreting complex visual semantics, such as gestures, body actions, or symbolic cues. Here, the semantic meaning associated with an action is replaced with that of another action of the same type. For instance, as shown in Fig~\ref{fig:motivation}, changing the meaning of an “OK” gesture to “rejection” or “sarcasm.”

\textbf{Entity Knowledge Conflict} centers on factual discrepancies surrounding entity attributes like nationality, occupation, or birth year. This is simulated by substituting the tail entity in a factual triple with another entity of the same type. For example, as shown in Fig~\ref{fig:motivation}, altering Elon Musk’s birth year from the correct value to 1974 or 1979.

% \textbf{Chart-based conflict}
% This type focuses on factual knowledge conflict, but visualized with a chart format, encompassing bar charts and pie charts. In realistic scenarios, one piece of knowledge could be represented in various visualization formats, and different visualizations could convey inconsistent information.  To construct such conflict, we propose first replacing the original factual knowledge with counterfactual content and then plotting a bar or pie figures to visualize this knowledge. As shown in Fig~\ref{fig:motivation}, the deepest lake is changed to Tanganyika or the Caspian Sea.

% \textbf{Chart-based conflict} targets factual inconsistencies represented through visualizations such as bar or pie charts. In this case, original factual statements are replaced with counterfactual content, which is then visualized using misleading charts. For example, representing Tanganyika or the Caspian Sea as the world’s deepest lake instead of the correct answer Baikal.

% \vspace{-3mm}
\section{MMKC-Bench}
% \vspace{-3mm}
% In this section, we describe the pipeline of leveraging various knowledge sources to develop MMKC-Bench, a QA based benchmark consisting of 1,881 instances that cover four kinds of multimodal knowledge conflicts, as shown in Fig~\ref{fig:pipline}.

\begin{figure}[t!]
 \vspace{-3mm}
  \centering
\includegraphics[width=1.0\linewidth]{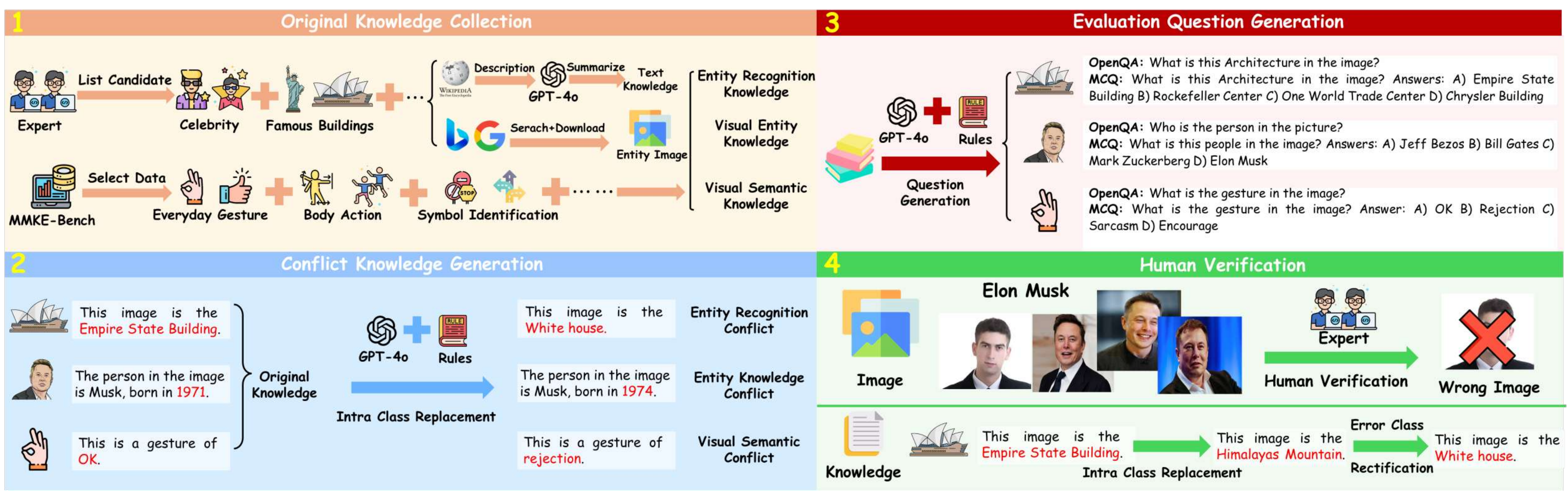}
  \caption{The construction pipeline of MMKC-Bench.}
  \label{fig:pipline}
  \vspace{-6mm}
\end{figure}

In this section, we present the pipeline for constructing MMKC-Bench, a QA-based benchmark comprising 1,573 instances that encompass three types of multimodal knowledge conflicts, as illustrated in Fig~\ref{fig:pipline}.

\vspace{-3mm}
\subsection{Original Knowledge Collection}
% including text + images
% We begin with original knowledge, we first list candidate fine-grained entity types, visual semantics, or pieces of knowledge, and then collect or draw their corresponding images and descriptions.

We begin with original knowledge collection by first listing candidate entity types or visual semantics. Then, we collect  the corresponding images and descriptions.

\begin{wrapfigure}{r} 
{0.43\textwidth} % r: 右侧, l: 左侧
    % \vspace{-3mm}
    \centering
\includegraphics[width=0.42\textwidth]{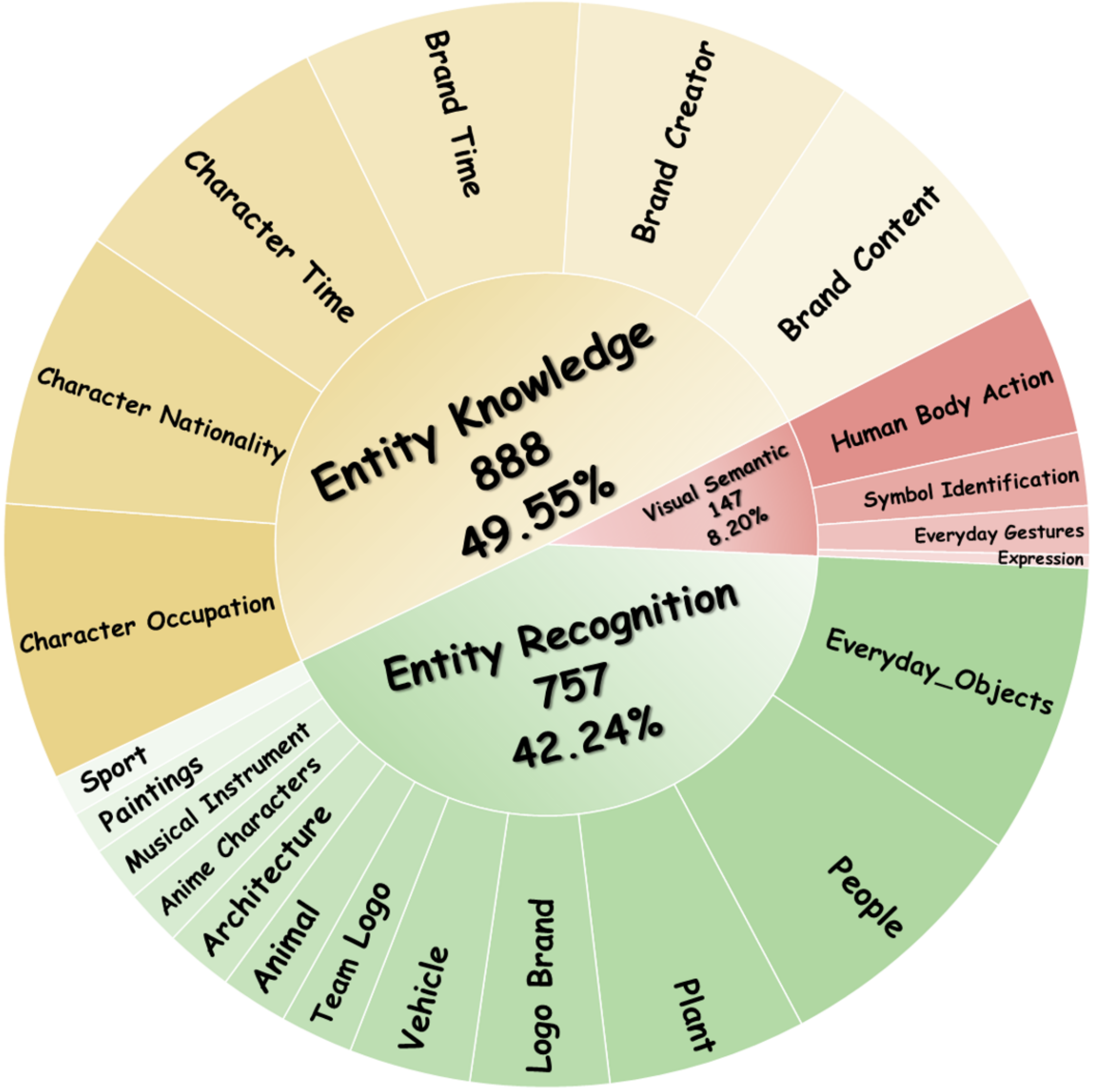}
    \caption{The data types of MMKC-Bench.}
        \label{fig:type}
    \vspace{-3mm}
\end{wrapfigure}
For entity recognition and entity knowledge conflicts, we manually define multiple candidate visual entity types (e.g., person, building). For each type, we use a large language model (LLM) to generate a list of the most prominent entities (e.g., Messi under the “person” category). Once the entity list is obtained, we crawl their images from Google and retrieve entity descriptions from Wikipedia summary dumps\footnote{\url{[https://dumps.wikimedia.org/}}, which are then summarized by an LLM to retain essential information. For visual semantic conflicts, we follow prior work on knowledge editing with visual semantic modifications~\cite{yuntaodu}, focusing on four categories of visual semantic knowledge: everyday gestures, human body actions, human emotions, and symbol identification. Since these types are already included in MMKC-Bench, we directly obtain the original visual semantic knowledge, consisting of paired images and their meanings, from the MMKC-Bench dataset, rather than collecting them manually.

\vspace{-3mm}
\subsection{Conflict Knowledge Generation}
\vspace{-2mm}

Considering the multimodal nature of LMMs, we generate conflict knowledge by deliberately introducing misalignments across modalities, with the assistance of large language models (LLMs).

For entity recognition, visual semantic, and entity knowledge conflicts, we retain the original image while modifying the textual component. Specifically, we replace the entity name, the meaning of the action, or the tail entity in a factual triple with another instance of the same type. For example, as illustrated in Fig~\ref{fig:motivation}, the name ``Empire State Building” is replaced with ``Eiffel Tower” or ``White House”, and Elon Musk’s birth year is altered to 1974 or 1979.

To simulate both context-memory conflicts and inter-context conflicts, we generate two conflicting versions for each piece of original knowledge following the above procedures. In the context-memory setting, one conflicting version is randomly selected as internal evidence. In the inter-context setting, both conflicting versions are provided as internal evidence.

\vspace{-3mm}
\subsection{Evaluation Question Generation}
\vspace{-2mm}

We adopt a visual question answering (VQA) format to construct evaluation questions and answers, leveraging LLMs for automatic generation. The specific prompts used are provided in the appendix.

We consider two types of questions: multiple-choice questions (MCQs) and open-ended questions (OQAs). For entity recognition and semantic recognition conflicts, the questions focus on identifying the entity name or interpreting the semantic meaning depicted in the image. For entity knowledge conflicts, the questions target fine-grained factual knowledge, such as querying a person’s occupation or age. In MCQs, each question includes four answer options: one answer within the models' internal knowledge, two answers from the conflicting knowledge variants, and one unrelated distractor option.

% adhere to four key evaluation principles to generate both the questions and answers. The reliability
% and portability questions are generated by prompting LLM and we show the prompts in the appendix.

\vspace{-3mm}
\subsection{Human Verification and Benchmark Statistics}
\vspace{-2mm}

% During benchmark construction, we manually collected, reviewed, and filtered the samples multiple times. In the original knowledge collection stage, we conducted a manual review of the images associated with each entity, semantic action and knowledge to ensure the quality of the collected images. 
% Furthermore, after counterfactual editing and question generation, we manually reviewed and filtered the questions, revised unsuitable questions, and corrected wrong answers.  We believe that this human effort contributes to the creation of a high-quality evaluation benchmark for the community.

During the construction of the benchmark, we conducted multiple rounds of manual collection, review, and filtering to ensure data quality. In the original knowledge collection stage, all images associated with each entity, semantic action, and piece of knowledge were manually reviewed to ensure their accuracy and relevance. Furthermore, following counterfactual editing and question generation, we performed additional manual verification, filtering out inappropriate samples, revising ambiguous or ill-formed questions, and correcting incorrect answers. We believe that this extensive human effort has been instrumental in ensuring the reliability and quality of the benchmark. 
% ultimately contributing a valuable resource to the research community.

\begin{wraptable}{r}{0.47\textwidth}
\vspace{-4mm}
\centering % 这会使得整个表格水平居中
\caption{The statistics of MMKC-Bench.}
\label{tab:statis}
\resizebox{0.97\linewidth}{!}{
\begin{tabular}{l|c|c|c}
\toprule
 & \textbf{\#Types} & \textbf{\#Instances} &  \textbf{\#Images} \\
\midrule
\textbf{{Visual Entity Conflict}}     &  13  & 757 & 2,271 \\
\textbf{Entity Knowledge Conflict}     & 6  & 669 & 669  \\
\textbf{Visual Semantic Conflict}    &  4 & 147 &441  \\
\bottomrule
\end{tabular}
}
% \vspace{-3mm}
\end{wraptable}
The statistical information of MMKC-Bench is shown in Tab~\ref{tab:statis}. As we can see, MMKC-Bench encompasses three types of conflict knowledge, containing 1,573 pieces of knowledge and 3,381 images. These knowledge spans 23 fine-grained types, highlighting the diversity of MMKC-Bench. 

\section{Experiment}
% \vspace{-3mm}
\subsection{Setup}
% \vspace{-2mm}
\textbf{Models} In multimodal knowledge conflict scenarios, the model input consists of multiple interleaved images and texts. Therefore, we select LMMs that perform well in multi-image understanding. Specifically, we conduct a comprehensive evaluation on 9 LMMs across 3 model series, with sizes ranging from 3B to 72B. The selected models include: \textbf{Qwen2.5-VL} (3B, 7B, 32B, 72B)~\cite{bai2025qwen2}, \textbf{InternVL3} (8B, 14B, 38B, 78B)~\cite{zhu2025internvl3}, and \textbf{GPT-4o mini}~\cite{hurst2024gpt}.

\textbf{Settings} We consider two conflict-related tasks: conflict behavior analysis and conflict detection. The former investigates how models behave under conflicting scenarios, while the latter evaluates whether models can correctly detect the presence of conflict.

For conflict behavior analysis, we consider two types of conflict scenarios: context-memory conflict and inter-context conflict. In context-memory conflict, one piece of conflicting external evidence, composed of an image and associated text, is provided as an in-context example. In inter-context conflict, two conflicting pieces of evidence about the same knowledge are provided as in-context examples. The model is then required to answer an evaluation question based on this context.

For conflict detection, we explore both coarse-grained and fine-grained conflict detection. In the coarse setting, a full piece of evidence (either conflicting or non-conflicting) is provided in context, and the model must determine whether a conflict exists by answering “yes” or “no”. Following previous work~\cite{wang2023resolving}, the fine-grained setting involves providing only one single sentence, which is the subset of full evidence, and the model must again judge whether a conflict is present.

\textbf{Evaluation Metrics}  
For conflict behavior analysis, we assess how conflicting contexts influence the model’s answers to QA pairs. Each model prediction under a conflict scenario is categorized into one of three types: (1) consistent with the model’s answer in the non-conflict setting, (2) consistent with the external conflicting evidence, and (3) inconsistent with both, referred to as an irrelevant answer. To enable this, we first perform QA under a non-conflict setting to establish the model’s internal knowledge. We then compute three ratios: Original Answer Ratio (OAR), Counter Answer Ratio (CAR), and Irrelevant Answer Ratio (IAO), with OAR + CAR + IAO = 1 across the dataset.

For conflict detection, we treat this as a binary classification task. If a knowledge conflict exists, the model should output ``yes”; otherwise, it should output ``no”, in both coarse and fine-grained settings. Accordingly, we report the detection accuracy as the evaluation metric.

% \newpage

\begin{table}[t]
\vspace{-3mm}
 \centering
    \caption{
    % Results of both context-memory conflict and inter-context on MMKC-Bench with multi-choice question format. 
    Results of both context-memory and inter-context conflicts on MMKC-Bench using the multiple-choice question format.
    }
    \label{main_res_mcq}
 \resizebox{0.85\textwidth}{!}{ %  
\begin{tabular}{c|cccc|cccc|cccc}
\toprule
    & \multicolumn{4}{c}{\textbf{Qwen2.5-VL-7B}}    & \multicolumn{4}{c}{\textbf{InternVL3-8B}} &     \multicolumn{4}{c}{\textbf{GPT-4o mini}}     \\
    \midrule
    & ER                & EK              & VS    & Avg.  & ER              & EK             & VS   & Avg.  & ER             & EK            & VS   & Avg.  \\
    \midrule
    & \multicolumn{12}{c}{Context-Memory Conflict}    \\
    \midrule
	\textbf{OAR} & 0.81              & 0.50            & 0.71  & 0.67 & 0.48            & 0.46           & 0.70 & 0.49 & 0.78           & 0.54          & 0.61 & 0.66 \\
	\textbf{CAR} & 0.15              & 0.48            & 0.21  & 0.30 & 0.41            & 0.45           & 0.26 & 0.41 & 0.11           & 0.43          & 0.30 & 0.27 \\
	\textbf{IAR} & 0.04              & 0.02            & 0.08  & 0.03 & 0.11            & 0.09           & 0.04 & 0.09 & 0.11           & 0.03          & 0.09 & 0.07 \\
    \midrule
    & \multicolumn{12}{c}{Inter-Context Conflict}      \\
    \midrule
	\textbf{OAR} & 0.87              & 0.51            & 0.72  & 0.70 & 0.41            & 0.47           & 0.66 & 0.46 & 0.76           & 0.47          & 0.57 & 0.62 \\
	\textbf{CAR} & 0.10              & 0.48            & 0.25  & 0.28 & 0.53            & 0.51           & 0.30 & 0.50 & 0.19           & 0.50          & 0.40 & 0.34 \\
	\textbf{IAR} & 0.03              & 0.01            & 0.03  & 0.02 & 0.07            & 0.02           & 0.05 & 0.04 & 0.05           & 0.03          & 0.03 & 0.04 \\
\bottomrule
\end{tabular}
}
% \vspace{-3mm}
\end{table}

\begin{table}[t]
 \centering
    \caption{
    % Results of both context-memory conflict and inter-context on MMKC-Bench with open question answer format. 
    Results of both context-memory and inter-context conflicts on MMKC-Bench using the open-ended question answering format.
    }
    \label{main_res_oqa}
    \resizebox{0.85\textwidth}{!}{ %  
\begin{tabular}{c|cccc|cccc|cccc}
\toprule
    & \multicolumn{4}{c|}{\textbf{Qwen2.5-VL-7B}} & \multicolumn{4}{c|}{\textbf{InternVL3-8B}} & \multicolumn{4}{c}{\textbf{GPT-4o mini}} \\
    \midrule
    & ER     & EK     & VS     &  Avg.     & ER     & EK     & VS     & Avg.   & ER     & EK    & VS    & Avg.   \\
    \midrule
    & \multicolumn{12}{c}{Context-Memory Conflict}                    \\
  \midrule  
\textbf{OAR} & 0.66   & 0.26   & 0.13   & 0.44  & 0.60   & 0.27   & 0.27   & 0.43  & 0.76   & 0.36  & 0.45  & 0.56  \\
\textbf{CAR} & 0.20   & 0.62   & 0.48   & 0.40  & 0.19   & 0.53   & 0.15   & 0.33  & 0.02   & 0.47  & 0.05  & 0.21  \\
\textbf{IAR} & 0.14   & 0.10   & 0.39   & 0.15  & 0.22   & 0.20   & 0.58   & 0.24  & 0.22   & 0.17  & 0.50  & 0.22  \\
    \midrule
    & \multicolumn{12}{c}{Inter-Context Conflict}                  \\
    \midrule
\textbf{OAR} & 0.65   & 0.14   & 0.05   & 0.38  & 0.46   & 0.15   & 0.18   & 0.30  & 0.82   & 0.37  & 0.37  & 0.58  \\
\textbf{CAR} & 0.21   & 0.76   & 0.77   & 0.50  & 0.40   & 0.72   & 0.52   & 0.54  & 0.06   & 0.47  & 0.07  & 0.24  \\
\textbf{IAR} & 0.14   & 0.10   & 0.18   & 0.12  & 0.14   & 0.14   & 0.30   & 0.15  & 0.12   & 0.16  & 0.56  & 0.18  \\
\bottomrule
\end{tabular}
}
% \vspace{-3mm}
\end{table}

\subsection{Model Behavior Analysis}

\begin{figure}[t!]
   % \vspace{-3mm}
   \centering \includegraphics[width=1.0\linewidth]{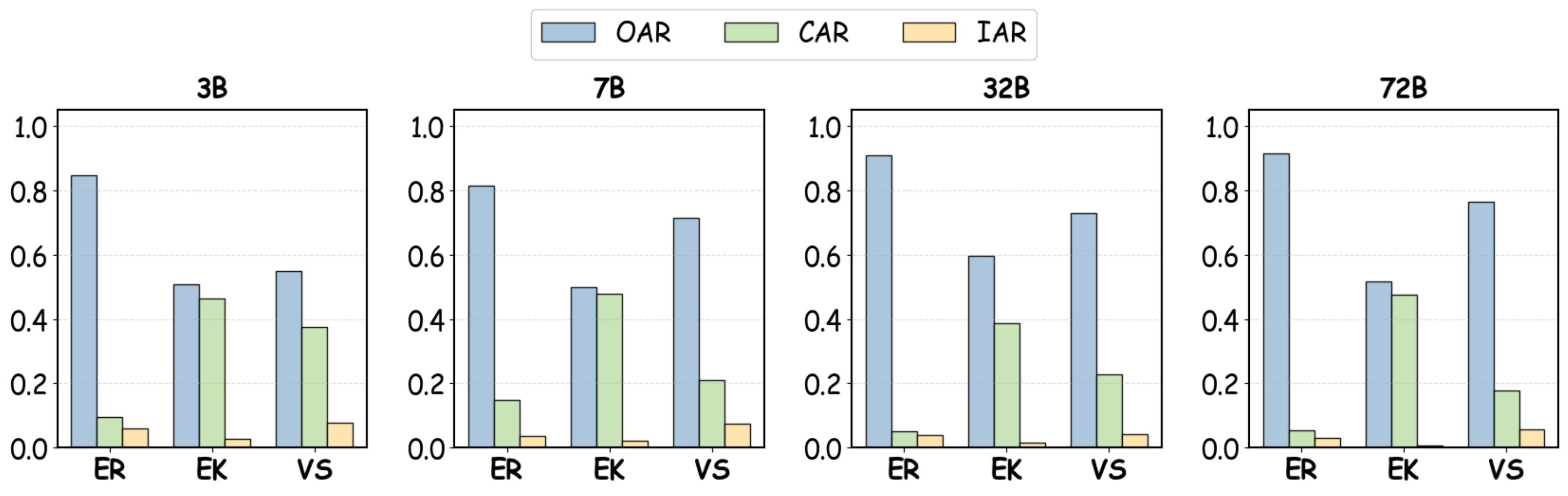}
  \caption{The results of Qwen2.5-VL with different model sizes under context-memory conflict with multi-choice question format.}
\label{fig:model_size_qwenvl_ie_mcq}
   \vspace{-3mm}
 \end{figure}

\begin{figure}[t!]
   \centering
\includegraphics[width=1.0\linewidth]{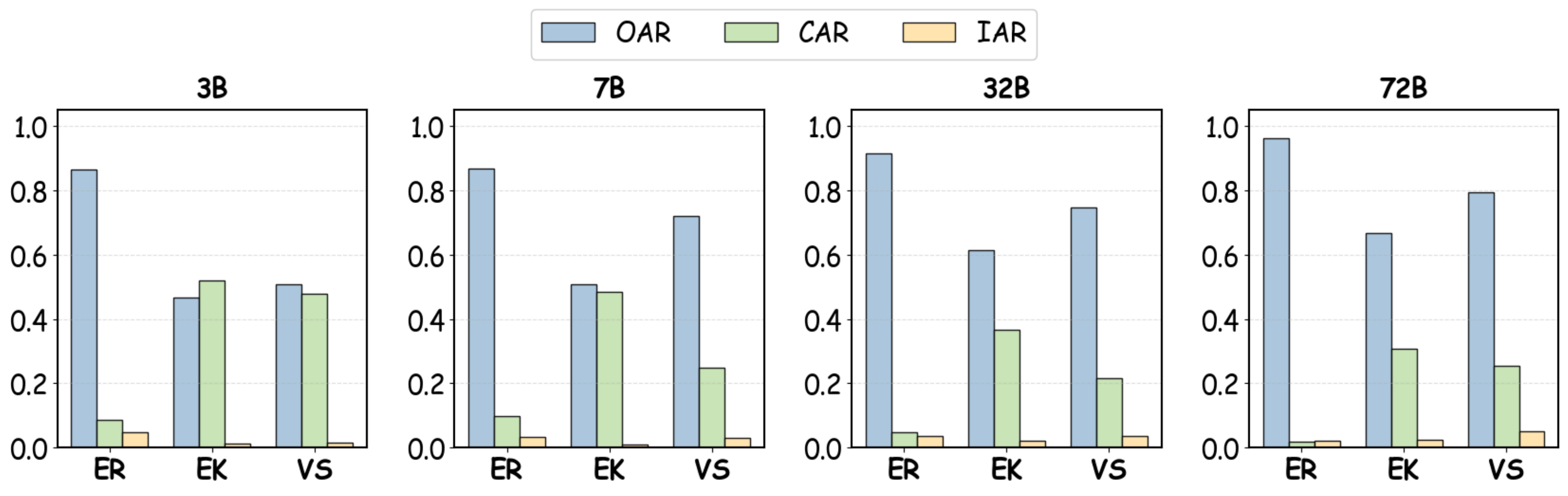}
  \caption{ The results of Qwen2.5-VL with different model sizes under inter-context conflict with multi-choice question format.}
\label{fig:model_size_qwenvl_ee_mcq}
   \vspace{-3mm}
 \end{figure}

The results under both context-memory and inter-context conflict scenarios, using multiple-choice and open-ended question formats, are presented in Table~\ref{main_res_mcq}, Table~\ref{main_res_oqa}, Fig.~\ref{fig:model_size_qwenvl_ie_mcq}, and Fig.~\ref{fig:model_size_qwenvl_ee_mcq}. Based on these results, we draw the following observations:

%%%%%%%%%%%
 \textbf{1) LMMs are more receptive to internal knowledge than to external evidence.}
As shown in Table~\ref{main_res_mcq} and Table~\ref{main_res_oqa}, under context-memory conflicts, the average OAR exceeds CAR in all cases (6 out of 6), indicating that LMMs tend to favor internal knowledge. Closed-source GPT-4o mini shows consistent results with open-source models, suggesting that even advanced closed models are insensitive to external evidence. This differs from LLMs, which have shown high receptiveness to external knowledge~\cite{su2024texttt,hou2024wikicontradict}. One reason for this contrast is the difference in training data formats: LLMs are typically trained on long text contexts involving multiple information sources, while LMMs are mostly trained on isolated image-text pairs. This limits their exposure to multi-source contexts and reduces their ability to integrate external information during inference.

This finding is important for designing multimodal RAG systems, as it reveals that LMMs may not naturally leverage retrieved evidence and instead rely on parametric knowledge. Thus, improving LMMs’ ability to incorporate external information is important, which may require innovations in training paradigms and model architecture.

\textbf{2) LMMs are more sensitive to knowledge-related conflicts and less sensitive to recognition-based conflicts.}
We group the three conflict types into recognition-based (entity recognition, visual semantics) and knowledge-related (entity knowledge). LMMs show lower OARs on knowledge-related conflicts than of recognition-based conflicts, indicating greater sensitivity to factual inconsistencies.  For example, entity recognition conflicts yield an OAR as low as 0.26 on Qwen2.5-VL-7B. While entity recognition conflicts often show the highest OARs, suggesting LMMs more easily rely on internal knowledge for perception tasks.

As shown in prior work~\cite{shao2024cognition,chen2024pca}, perception and cognition are core abilities of LMMs. Recognition-based tasks rely on visual-text alignment (perception), whereas knowledge-related tasks involve cognitive reasoning over facts. LMMs are mainly trained on perception tasks like VQA, grounding, and captioning, with less exposure to cognitively demanding data. This imbalance results in stronger perception than reasoning abilities. Therefore, LMMs lean on internal memory for recognition, but may turn to external sources for knowledge-intensive tasks. This highlights the need to enrich training data with cognitively challenging examples to strengthen LMM reasoning capabilities.

\textbf{3) When provided with more external evidence, LMMs exhibit greater alignment with external information, though the improvement remains limited.}
Compared to context-memory conflict scenarios, models generally achieve higher CARs under inter-context conflicts, suggesting a slight increase in reliance on external evidence. This is because, given more internal information, the model output would be affected more. However, the overall improvement is limited: the largest increase in CAR is 21\% on average, while the smallest average improvement is only about -2\%. These results reaffirm that LMMs predominantly rely on their internal parametric knowledge, even when presented with multiple external sources.

\textbf{4) Larger models exhibit a stronger promoting effect across all conflict types.}
As illustrated in Fig.\ref{fig:model_size_qwenvl_ie_mcq} and Fig.\ref{fig:model_size_qwenvl_ee_mcq}, the Overall Agreement Rate (OAR) generally increases with model size within the Qwen2.5-VL series. Specifically, the OAR improves progressively as the model scales from 3B to 7B, 13B, and 70B, reflecting gains across entity recognition conflict, entity knowledge conflict, and visual semantic conflict. This trend suggests that larger models are more strongly influenced by their internal knowledge.  This enhanced capability may stem from exposure to more extensive training data, enabling larger models to develop stronger mechanisms for resolving conflicts.

\textbf{5) While performance differs between the two question formats, the overall trends remain consistent.}
Under the two question formats, the models exhibit different performance levels. For instance, in the open-ended question format, models tend to achieve higher IAR, suggesting that the open-ended nature of the task introduces greater variability in the model outputs. Despite these differences in absolute performance, the overall trend across both formats remains consistent, demonstrating the robustness of the proposed benchmark across varying evaluation settings.

% \newpage

\begin{table}[t]
 \centering
 \LARGE
 \renewcommand{\arraystretch}{1.2}
    \caption{Results of coarse-grained and fine-grained conflict detection on MMKC-Bench.
    }
    \label{main_det}
    \resizebox{1.0\textwidth}{!}{ %  
\begin{tabular}{l|ccc|ccc|ccc|c|ccc}
\toprule
& \multicolumn{10}{c|}{\textbf{Coarse-Grained Detection}} & \multicolumn{3}{c} {\textbf{Fine-Grained Detection}} \\
\midrule
            & \multicolumn{3}{c|}{ER}         & \multicolumn{3}{c|}{EK}         & \multicolumn{3}{c|}{VS}          & Avg.  & \multicolumn{3}{c}{EK}         \\
            \midrule
            & Non-Conflict & Conflict & Avg.  & Non-Conflict & Conflict & Avg.  & Non-Conflict & Conflict & Avg.    & & Non-conflict & Conflict & Avg.  \\
            \midrule
\textbf{Qwen2.5-VL-7B}   & 0.92   & 0.87 & 0.89 & 0.89    & 0.51  & 0.70 & 0.67   & 0.89     & 0.78   & 0.79 & 0.76      & 0.65  & 0.71 \\
\textbf{InternVL3-8B} & 0.95         & 0.44     & 0.69 & 0.98         & 0.67    & 0.82 & 0.87         & 0.72     & 0.79  & 0.75 & 0.92         & 0.35     &0.64  \\
\textbf{GPT-4o mini} & 0.73         & 0.88     & 0.80 & 0.66         & 0.76    & 0.71 & 0.63        & 0.82     & 0.73  & 0.76 & 0.82        & 0.61     & 0.72 \\
\bottomrule
\end{tabular}
}
\vspace{-3mm}
\end{table}

%% Fine-grained type.
% \vspace{-3mm}
\subsection{Conflict  Detection Analysis}
% \vspace{-2mm}

The results of both coarse-grained and fine-grained conflict detection are shown in Table~\ref{main_det}. Based on the results, we have the following findings:

% \textbf{1. **}

% \textbf{1. LMMs could effectively determine whether there is knowledge conflict, and recognize conflicts better under conflict scenarios than non-conflict scenarios in most cases.} As shown in Table~\ref{main_det}, the average detection accuracy is 79\%, 75\%, and 76\% for Qwen2.5-VL-7B, InternVL-8B, and GPT-4o mini, respectively, indicating that the LMMs could effivetyle acknowledge the presence of knowledge conflict. Besides, in most case (5/9), the detection accuracy under non-conflict is higher than that of the conflict scenario, which shows that the model detects better when no knowledge conflict occurs.

\textbf{1) LMMs can effectively identify the presence of knowledge conflicts and generally perform better in recognizing conflicts under conflict scenarios than non-conflict scenarios.} As shown in Table~\ref{main_det}, the average detection accuracy reaches 79\%, 75\%, and 76\% for Qwen2.5-VL-7B, InternVL-8B, and GPT-4o mini, respectively, indicating that LMMs are capable of reliably detecting the existence of knowledge conflicts. Moreover, in most cases (5 out of 9), the detection accuracy under non-conflict scenarios is higher than that under conflict scenarios, suggesting that models tend to detect more accurately when no knowledge conflict is present. Besides, it is also found that open-source models have similar or even better performance than closed-source models, showing a smaller gap between these models.

% \textbf{2. LMMs could effectively determine whether there is knowledge conflict under both coarse-grained and fine-grained conflict scenarios.} The detection accuracy under fine-grained scenarios is similar to or lower than that of under coarse-grained scenarios. The results show that under both scenarios, the LMMs could recognize conflict effectively, while the performance is lower under the fine-grained scenario, which aligns with the findings in previous work~\cite{wang2023resolving}.

\textbf{2) LMMs can effectively identify knowledge conflicts in both coarse-grained and fine-grained scenarios.} The detection accuracy under fine-grained scenarios is comparable to or slightly lower than that under coarse-grained scenarios. These results indicate that LMMs are capable of recognizing knowledge conflicts across both levels of granularity. However, the slightly lower performance in fine-grained settings is consistent with observations in previous work~\cite{wang2023resolving}.

\vspace{-3mm}
\subsection{Case Study}
\vspace{-2mm}

\begin{figure}[h]
  \centering
\includegraphics[width=1.0\linewidth]{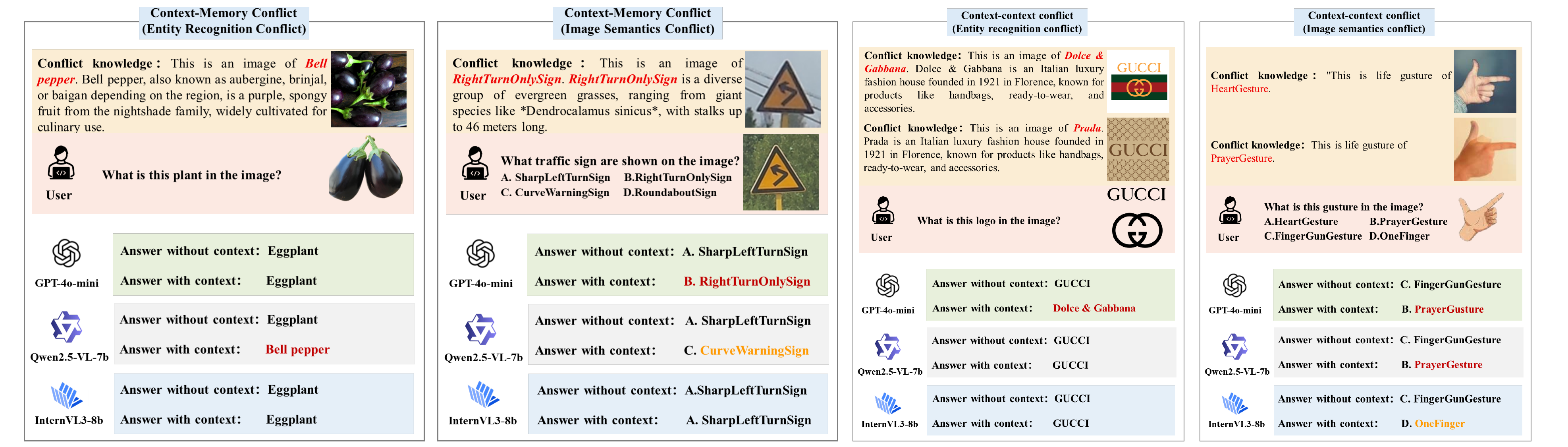}
  \caption{Case study of context-memory conflict and inter-context conflict involving entity recognition
and visual semantic conflicts.}
\label{fig:case_study}
  \vspace{-3mm}
\end{figure}

Two examples each of context-memory conflict and inter-context conflict involving entity recognition and visual semantic conflicts are presented in Fig.~\ref{fig:case_study}. The answers without context reflect the models' parametric (internal) knowledge, while the answers with context show the models' behavior when exposed to conflicting external evidence. As observed, across all models, the original answers align with the ground truth. However, under context-memory conflict scenarios, most models tend to rely on their internal knowledge, often ignoring the provided external evidence. When more external evidence is introduced in the inter-context setting, models are more likely to refer to external knowledge sources. Moreover, we also observe that under conflict scenarios, models may sometimes produce answers that are inconsistent with both internal knowledge and external evidence, such as ``curvewaringsign” in the second example.

\section{Conclusion}
% \vspace{-1mm}
In this paper, we propose MMKC-Bench, a  multimodal knowledge conflict benchmark aimed at analyzing factual conflicts in both context-memory and inter-context scenarios. Our benchmark covers three types of multimodal knowledge conflicts and incorporates two distinct conflict settings. Through extensive experiments on representative LMMs, we observe that most models are more receptive to internal (parametric) knowledge and exhibit limited sensitivity to external conflicting information. We hope our work will inspire further research in knowledge conflict resolution and the development of multimodal retrieval-augmented generation (RAG) frameworks. As for limitations, collecting real-world multimodal knowledge conflicts remains challenging. To address this, we employ counterfactual editing to synthesize conflict instances, which inevitably introduces a distribution gap between our benchmark and naturally occurring data. In the future, real-world multimodal knowledge conflict benchmarks, such as \cite{hou2024wikicontradict}, are needed to more accurately reflect real-world scenarios and enhance the robustness of model evaluations.

% \newpage

% \newpage

\bibliographystyle{unsrt}  
% \bibliography{reference}

\newpage

\renewcommand{\thesection}{\Alph{section}}
\section{DBENCHMARK CONSTRUCTION}
\subsection{ORIGINAL KNOWLEDGE COLLECTION}
In the process of collecting raw knowledge, we first select the most popular entities, then gather the corresponding raw knowledge of these entities using Wikipedia. Then, we collect the images of the entities to construct multimodal data, followed by building conflicting knowledge, and finally generating corresponding evaluation questions.

For entity recognition tasks, we first identify visually grounded entity categories (e.g., building, people) and then collect the most popular entities for each category using existing datasets or generated by LLMs. Next, we retrieve the raw description for each entity from Wikipedia. To ensure data quality, we apply an automated preliminary filtering process: excluding entities with raw knowledge shorter than 30 words and leveraging Wikipedia’s API to prioritize the most popular entities per category. Since raw knowledge may contain noise, logical inconsistencies, or fragmented text, we employ LLMs to summarize it while preserving the original information. Subsequently, we crawl images for each entity, manually inspect and remove low-quality visuals, and finally ensure a minimum of three images per entity.

For entity knowledge, we extended the entity recognition data and constructed standardized knowledge across two themes (individuals and brand logos) with three dimensions each. Specifically, for individuals, we compiled knowledge on birth dates, nationalities, and occupations; for brand logos, we organized data on establishment dates, founders, and primary products.

For visual semantic, we collected semantic knowledge across four categories: symbolic images, common gestures, body movements, and facial expressions. We sourced candidate instances from the MMKE dataset and constructed raw knowledge for each instance, primarily consisting of descriptive narratives about the image semantics. Additionally, we ensured that every instance was supported by at least three corresponding images retrieved from the MMKE dataset.

In summary, this benchmark encompasses a total of 1,573 pieces of knowledge and 3,381 images.

\begin{figure}
    \centering
    \includegraphics[width=0.65\linewidth]{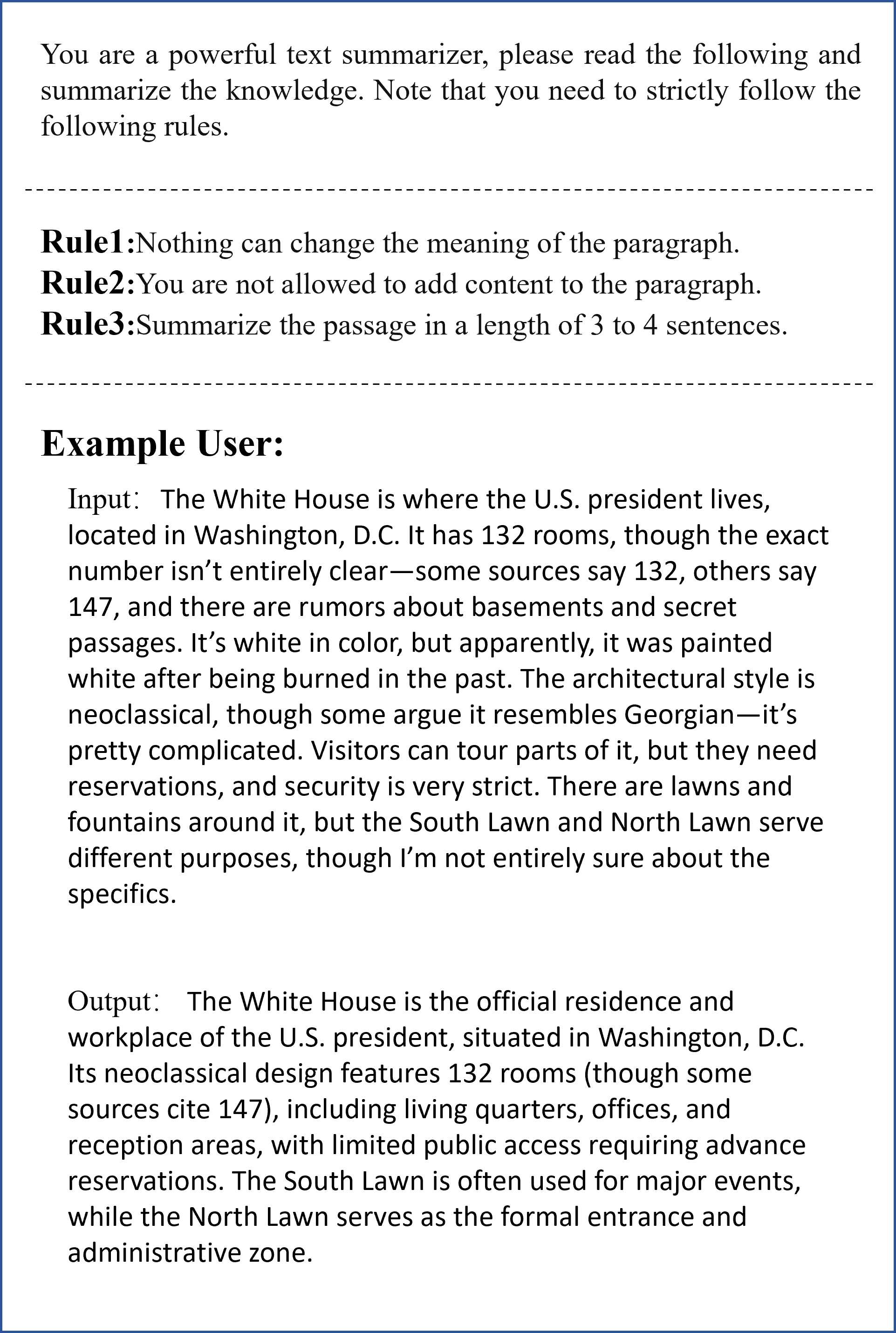}
    \caption{Prompt for summary of original knowledge}
    \label{fig:enter-label1}
    \vspace{-3mm}
\end{figure}

\begin{figure}
    \centering
    \includegraphics[width=0.65\linewidth]{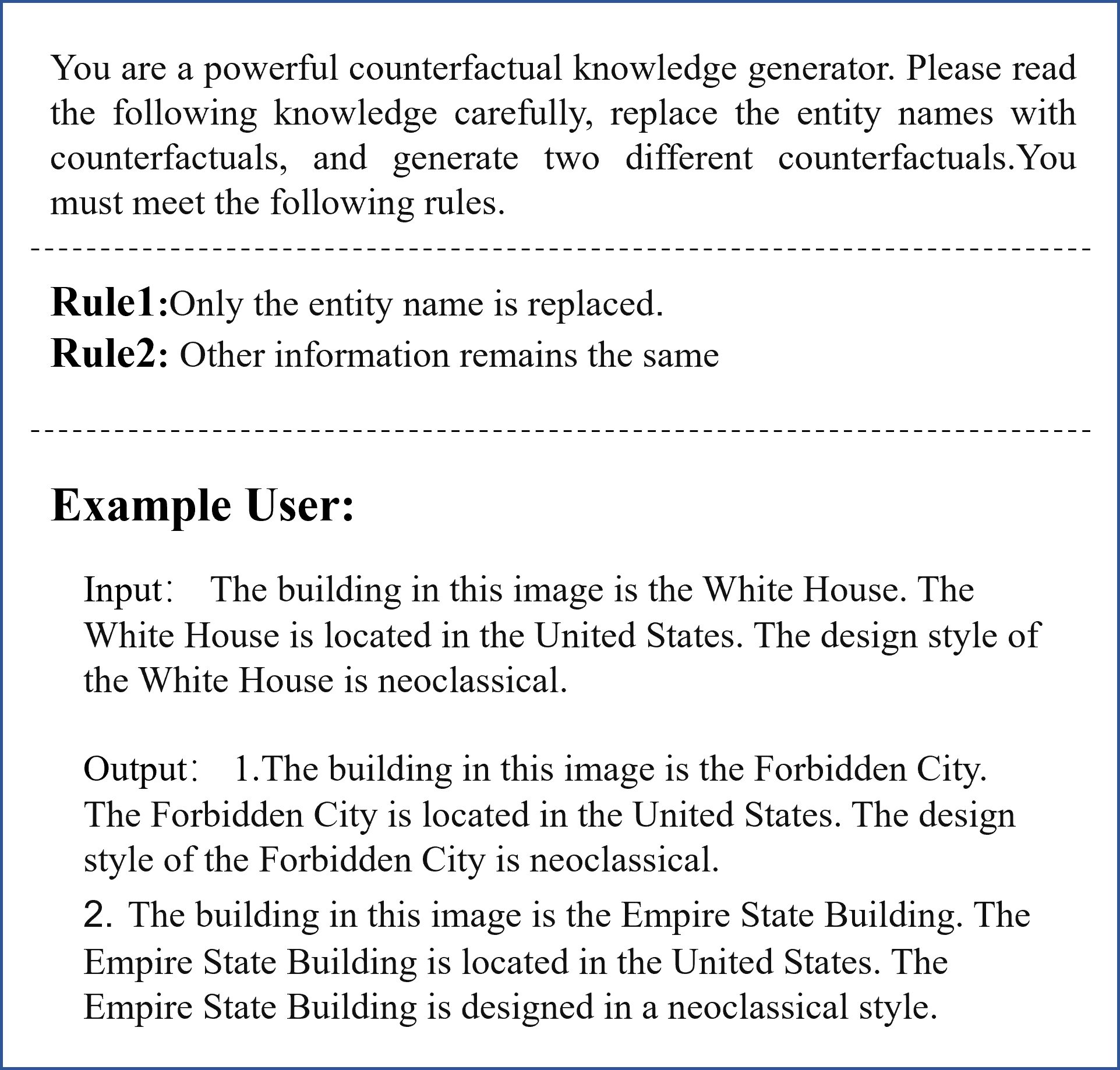}
    \caption{Prompt for Generate Entity Recognition conflicting knowledge}
    \label{fig:enter-label2}
\end{figure}

\begin{figure}
    \centering
    \includegraphics[width=0.65\linewidth]{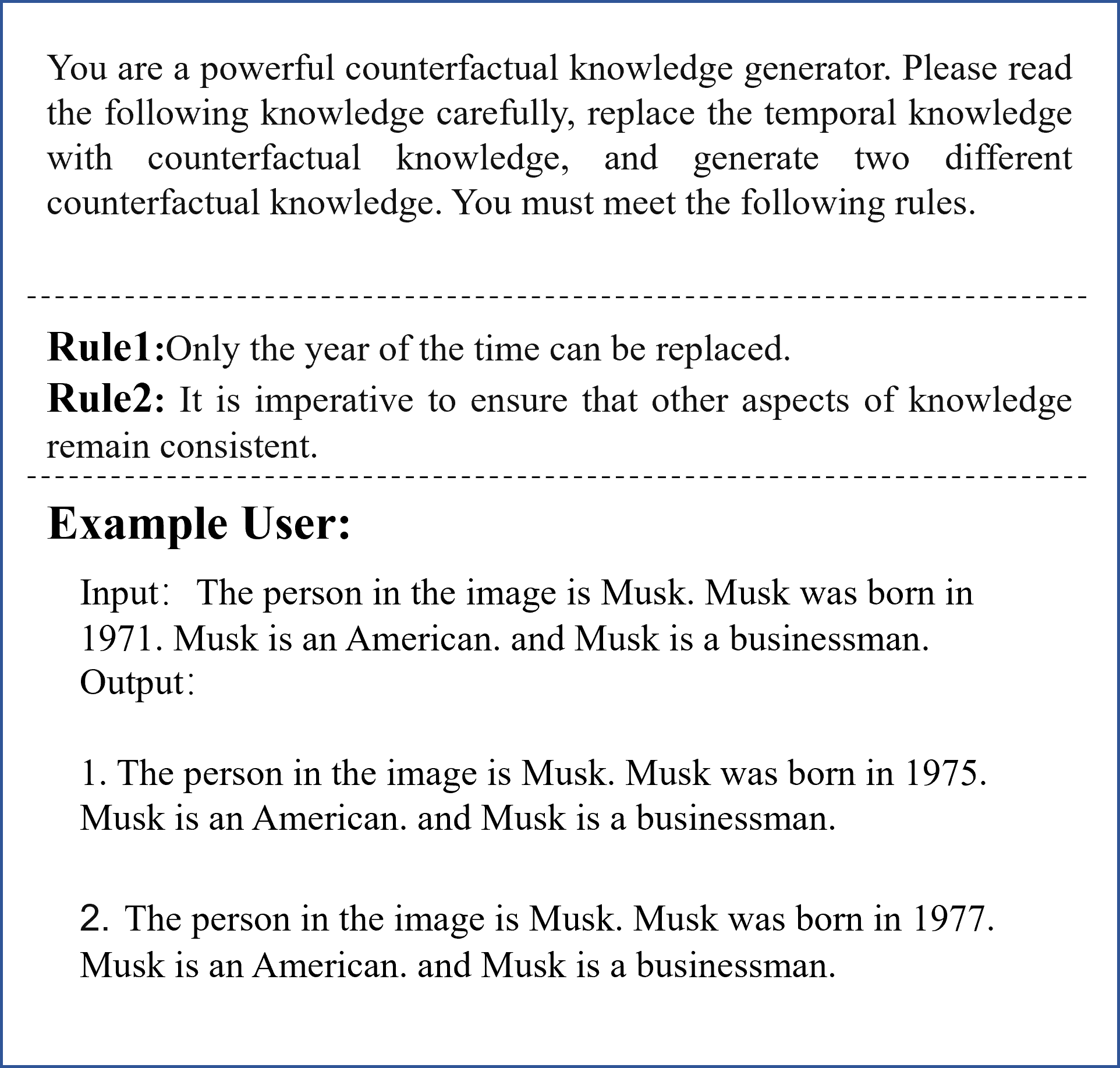}
    \caption{Prompt for Generate character time knowledge conflicting knowledge}
    \label{fig:enter-label3}
\end{figure}

\begin{figure}
    \centering
    \includegraphics[width=0.65\linewidth]{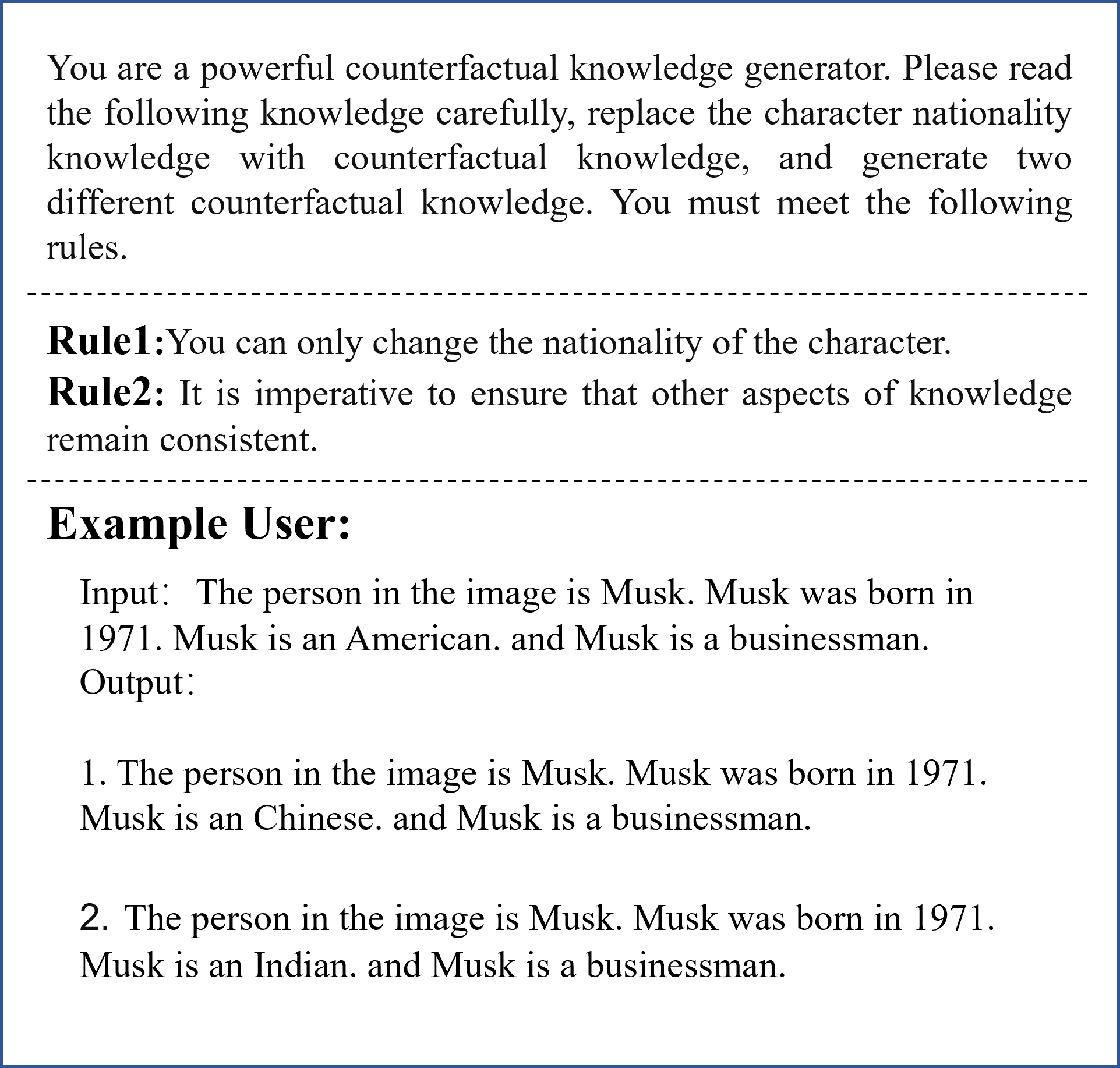}
    \caption{Prompt for Generate character country of citizenship knowledge conflicting knowledge}
    \label{fig:enter-label4}
\end{figure}

\begin{figure}
    \centering
    \includegraphics[width=0.65\linewidth]{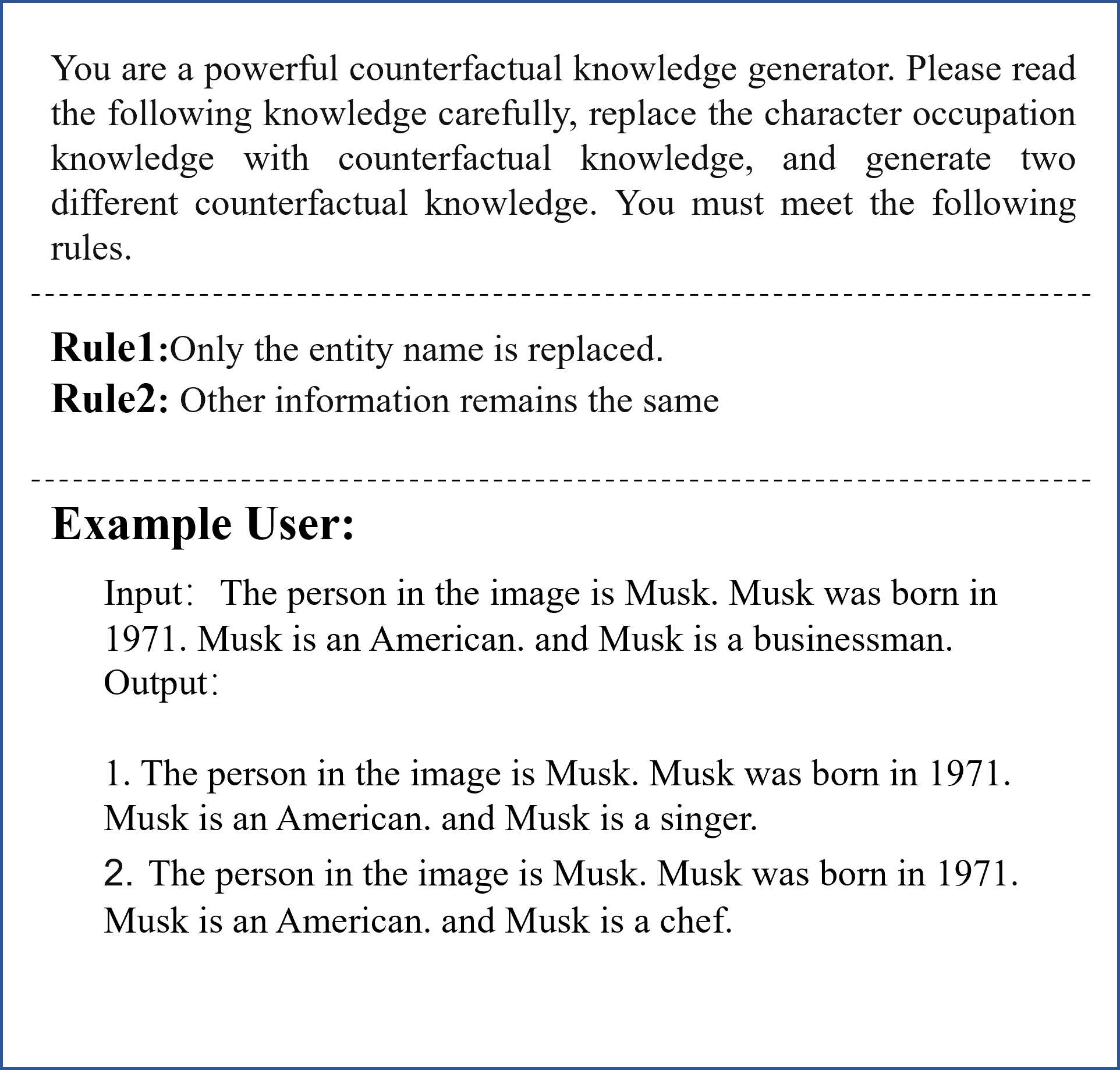}
    \caption{Prompt for Generate character occupation knowledge conflicting knowledge}
    \label{fig:enter-label5}
\end{figure}

\subsection{CONFLICT KNOWLEDGE GENERATION}
For conflict knowledge construction in entity recognition, we primarily leverage large language models to generate counterfactual conflict knowledge based on the original entity names while keeping all other content unchanged.

For the construction of conflicting knowledge in entity knowledge categories, we further developed multi-layered knowledge conflicts. Specifically, we employed large language models to generate counterfactual conflicting knowledge for each individual dimension while preserving all other dimensions unchanged.

For the construction of conflicting knowledge in image semantics, our focus lies on semantic vocabulary substitution. Specifically, we replace an entity or action with another of the same category—for example, substituting "happy" with "sad," or an "OK hand gesture" with a "phone hand gesture."

\section{LLM Prompts for Different Steps}
In this section, we provide a detailed list of all prompts for different steps, offering a clear reference
for understanding our experimental approach:

\begin{itemize}
    \item The prompt for summary and organization of original knowledge is shown in Figure\ref{fig:enter-label1}.
    \item The prompt for  generating conflicting Knowledge for Entity Recognition is shown in Figure\ref{fig:enter-label2}.
    \item The prompt for  generating conflicting Knowledge for character time is shown in Figure\ref{fig:enter-label3}.
    \item The prompt for generating conflicting Knowledge for character country of citizenship is shown in Figure\ref{fig:enter-label4}.
    \item The prompt for generating conflicting Knowledge for character occupation is shown in Figure\ref{fig:enter-label5}.
\end{itemize}

\section{EXPERIMENTS}
We experimented with the VLMEvalKit library, which uses PyTorch and integrates several large multimodal models. Experiments were conducted on NVIDIA L20 4BGB/A100 80GB GPUs.

\textbf{MLLMs.}To evaluate our benchmark, we conduct experiments on three representative MLLMs.

\begin{itemize}
    \item \textbf{Qwen2.5-VL:} Qwen2.5 VL is a multimodal large model launched by Alibaba. It achieves image-text joint understanding and generation by efficiently bridging the visual and language modalities. Its design continues the advantages of the Qwen series in the field of language models (LM), while combining visual encoders to form an end-to-end unified architecture.
    \item \textbf{InternVL3:} InternVL3 is the third-generation multimodal basic model launched by Shanghai AI Lab, focusing on the unification of general vision-language understanding and cross-modal generation capabilities. Its core goal is to achieve deep integration of images, videos, and texts through large-scale training and architecture innovation, and is suitable for open world scenarios.
    \item \textbf{GPT4o mini:} GPT4o mini is a lightweight language model designed by OpenAI for edge computing and low-cost deployment needs. While maintaining the core capabilities of GPT-4o, it achieves a balance between performance and efficiency through architecture compression and training optimization. It is suitable for scenarios such as real-time interaction and mobile integration.
\end{itemize}

\begin{figure}
    \centering
    \includegraphics[width=1\linewidth]{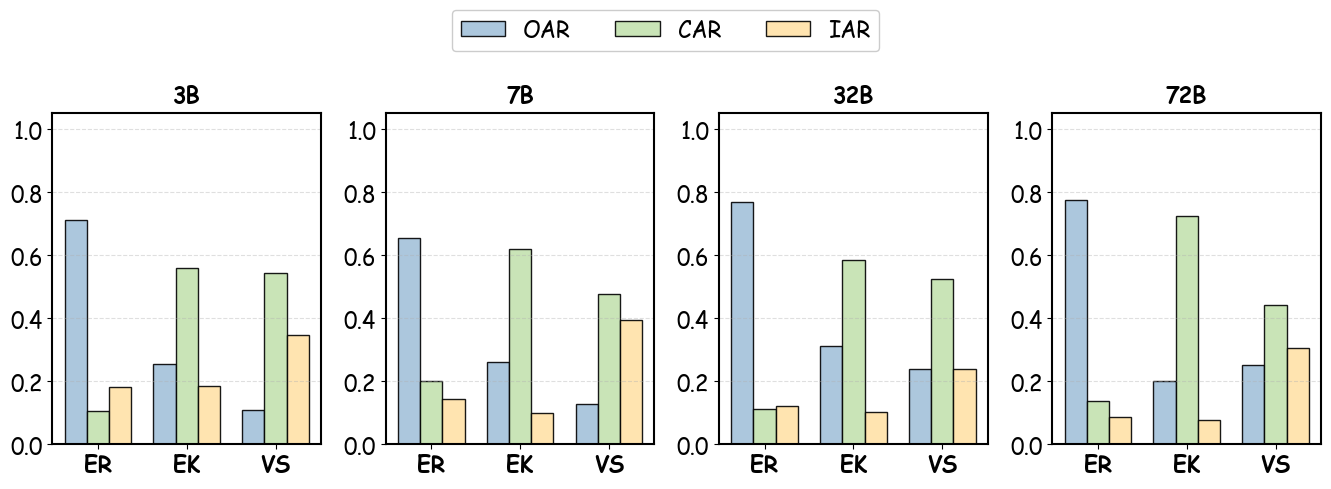}
    \caption{The results of Qwen2.5-VL with different model sizes under context-memory conflict with
open-ended question answering format}
    \label{qwenopenie}
\end{figure}

\begin{figure}
    \centering
    \includegraphics[width=1\linewidth]{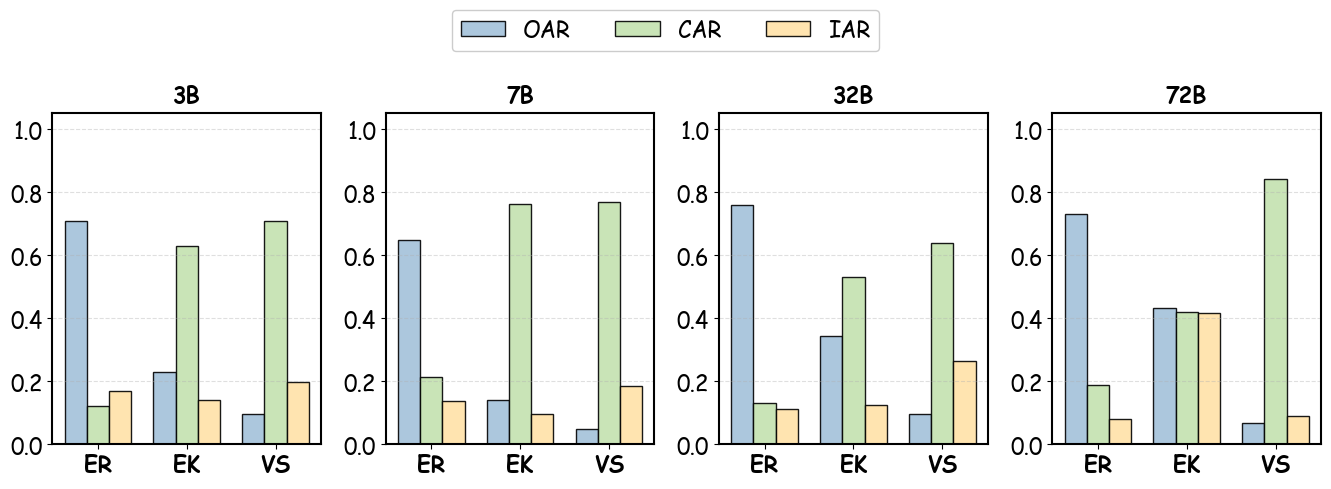}
    \caption{The results of Qwen2.5-VL with different model sizes under inter-context conflict with
open-ended question answering format}
    \label{qwenopenee}
\end{figure}

\begin{figure}
    \centering
    \includegraphics[width=1\linewidth]{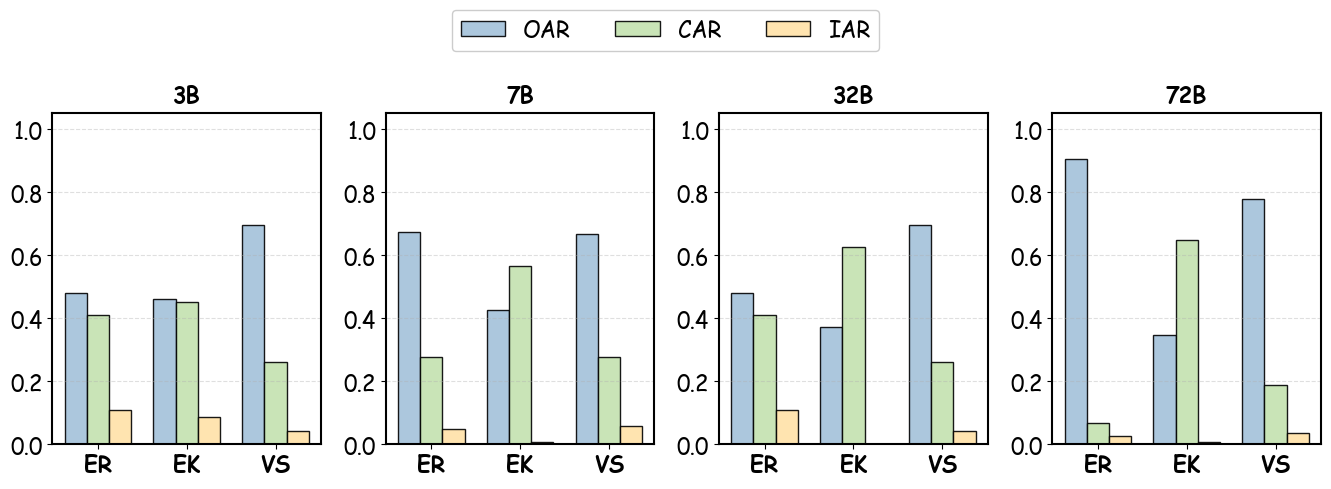}
    \caption{The results of InternVL3 with different model sizes under context-memory conflict with
 multiple-choice question format}
    \label{InternVL3mcqie}
\end{figure}

\begin{figure}
    \centering
    \includegraphics[width=1\linewidth]{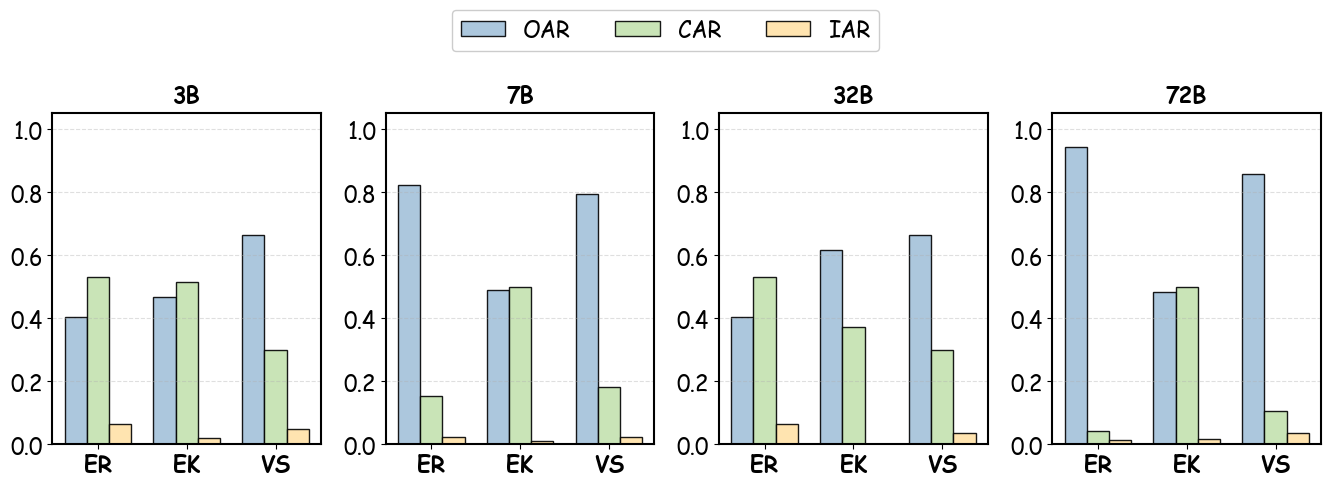}
    \caption{The results of InternVL3 with different model sizes under inter-context conflict with
open-ended question answering format}
    \label{InternVL3mcqee}
\end{figure}

\begin{figure}
    \centering
    \includegraphics[width=1\linewidth]{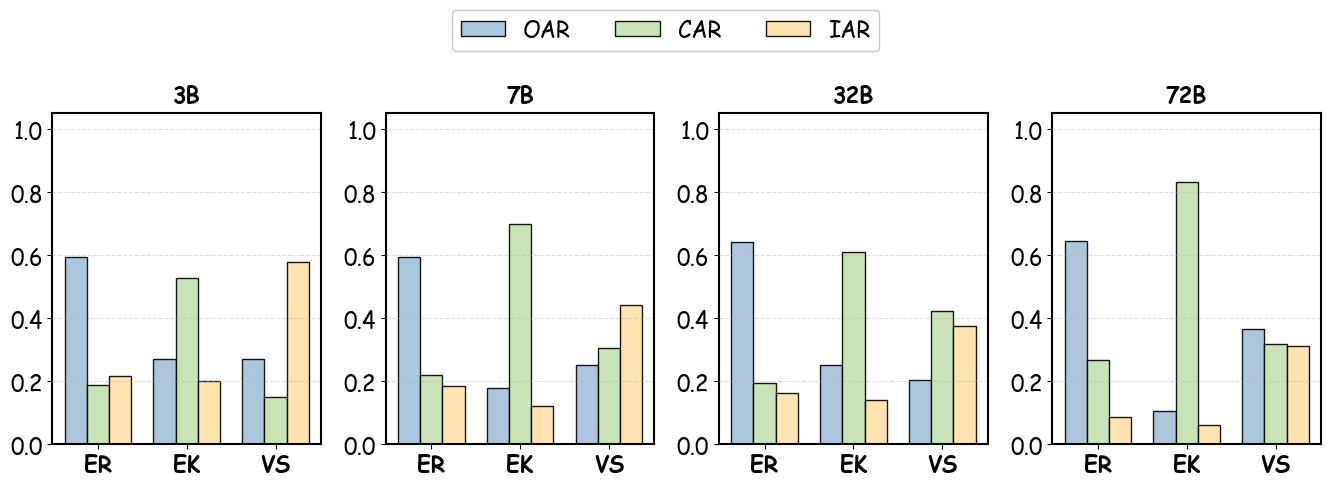}
    \caption{The results of InternVL3 with different model sizes under context-memory conflict with
open-ended question answering format}
    \label{InternVL3openie}
\end{figure}

\begin{figure}
    \centering
    \includegraphics[width=1\linewidth]{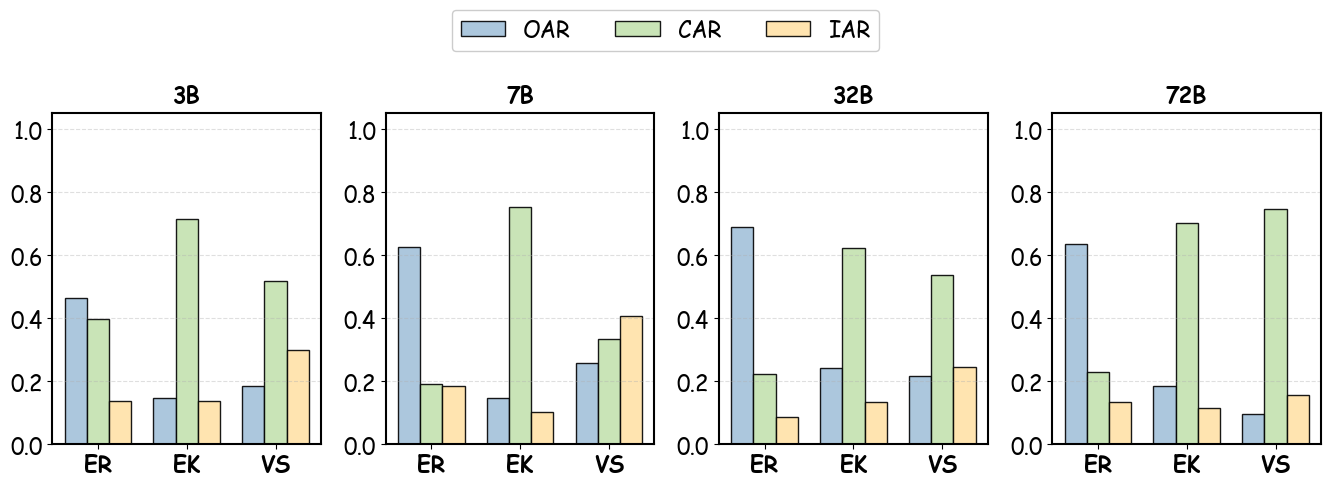}
    \caption{The results of InternVL3 with different model sizes under inter-context conflict with
open-ended question answering format}
    \label{InternVL3openee}
\end{figure}

\section{MORE RESULTS}
In this section, we present more experimental results from the main paper to provide more guidance.
\subsection{The results  with different model sizes}
% open-ended question answering format}

% We show the performance of the Qwen2.5-VL model series based on open-ended question answering under different model sizes under context-memory conflict and inter-context conflict in  Figure \ref{qwenopenie} and Figure\ref{qwenopenee}, respectively.
% The results of InternVL3 with different model sizes on the multiple-choice question format
% under context-memory conflict.  and inter-context conflict are shown in  Figure\ref{InternVL3mcqie} and Figure\ref{InternVL3mcqee}. 
% The results of InternVL3 with different model sizes with
% open-ended question answering format under context-memory conflict and inter-context conflict in Figure\ref{InternVL3openie} and Figure\ref{InternVL3openee}.

We present the performance of the Qwen2.5-VL model series on open-ended question answering under varying model sizes in the settings of context-memory conflict (Figure \ref{qwenopenie}) and inter-context conflict (Figure \ref{qwenopenee}). For InternVL3, we show results across different model sizes using the multiple-choice question format under context-memory conflict and inter-context conflict in Figure \ref{InternVL3mcqie} and Figure \ref{InternVL3mcqee}, respectively.
In addition, the performance of InternVL3 with the open-ended question format under both conflict settings is illustrated in Figure \ref{InternVL3openie} (context-memory conflict) and Figure \ref{InternVL3openee} (inter-context conflict). As we can see, larger models exhibit a stronger promoting effect across all conflict types. The Overall Agreement Rate (OAR) generally increases with model size within both the Qwen2.5-VL and InternVL3 series.  This trend suggests that larger models are more strongly influenced by their internal knowledge.  

\begin{table}[t]
\centering 
\caption{
Results of both context-memory and inter-context fine-grained Entity Knowledge conflicts on MMKC-Bench using open-ended question answering format.
}
\label{EKopen}
\resizebox{\textwidth}{!}{ % 自动调整宽度 
\begin{tabular}{c|*{6}{c}|*{6}{c}|*{6}{c}}
\toprule 
 & \multicolumn{6}{c|}{\textbf{Qwen2.5-VL-7B}} & \multicolumn{6}{c|}{\textbf{InternVL3-8B}} & \multicolumn{6}{c}{\textbf{GPT-4o mini}} \\
 \cmidrule(lr){2-7} \cmidrule(lr){8-13} \cmidrule(lr){14-19}
 & PT & PL & PC & LT & LC & LO & PT & PL & PC & LT & LC & LO & PT & PL & PC & LT & LC & LO \\
\midrule 
\multicolumn{19}{c}{\textbf{Context-Memory Conflict}} \\
\midrule 
\textbf{OAR} & 0.11 & 0.39 & 0.48 & 0.05 & 0.16 & 0.36 & 0.05 & 0.32 & 0.52 & 0.11 & 0.08 & 0.54 & 0.02 & 0.23 & 0.75 & 0.04 & 0.51 & 0.61 \\
\textbf{CAR} & 0.88 & 0.58 & 0.27 & 0.89 & 0.80 & 0.30 & 0.78 & 0.57 & 0.13 & 0.81 & 0.85 & 0.03 & 0.96 & 0.62 & 0.11 & 0.96 & 0.20 & 0.07 \\
\textbf{IAR} & 0.01 & 0.03& 0.25 & 0.05 & 0.04 & 0.34 & 0.17 & 0.11 & 0.34 & 0.08 & 0.07 & 0.43 & 0.02 & 0.14 & 0.13 & 0.00 & 0.28 & 0.32 \\
\midrule 
\multicolumn{19}{c}{\textbf{Inter-Context Conflict}} \\
\midrule 
\textbf{OAR} & 0.08 & 0.20 & 0.35 & 0.03 & 0.05 & 0.14 & 0.03 & 0.19 & 0.19 & 0.04 & 0.03 & 0.39 & 0.03 & 0.30 & 0.68 & 0.03 & 0.51 & 0.66 \\
\textbf{CAR} & 0.89 & 0.78 & 0.38 & 0.97 & 0.92 & 0.64 & 0.71 & 0.75 & 0.62 & 0.89 & 0.96 & 0.36 & 0.96 & 0.60 & 0.01 & 0.97 & 0.23 & 0.05 \\
\textbf{IAR} & 0.03 & 0.02 & 0.27 & 0.00 & 0.03 & 0.23 & 0.26 & 0.06 & 0.19 & 0.07 & 0.01 & 0.23 & 0.01 & 0.10 & 0.30 & 0.00 & 0.26 & 0.28 \\
\bottomrule 
\end{tabular}
}
\end{table}

\begin{table}[t]
\centering 
\caption{
Results of both context-memory and inter-context fine-grained Entity Knowledge conflicts on MMKC-Bench using the multiple-choice question format.
}
\label{EKMOQ}
\resizebox{\textwidth}{!}{ % 自动调整宽度 
\begin{tabular}{c|*{6}{c}|*{6}{c}|*{6}{c}}
\toprule 
 & \multicolumn{6}{c|}{\textbf{Qwen2.5-VL-7B}} & \multicolumn{6}{c|}{\textbf{InternVL3-8B}} & \multicolumn{6}{c}{\textbf{GPT-4o mini}} \\
 \cmidrule(lr){2-7} \cmidrule(lr){8-13} \cmidrule(lr){14-19}
 & PT & PL & PC & LT & LC & LO & PT & PL & PC & LT & LC & LO & PT & PL & PC & LT & LC & LO \\
\midrule 
\multicolumn{19}{c}{\textbf{Context-Memory Conflict}} \\
\midrule 
\textbf{OAR} & 0.04 & 0.26 & 0.93 & 0.03 & 0.77 & 0.96 & 0.04 & 0.26 & 0.85 & 0.03 & 0.72 & 0.86 & 0.27 & 0.28 & 0.96 & 0.23 & 0.67 & 0.81 \\
\textbf{CAR} & 0.96 & 0.71 & 0.05 & 0.91 & 0.20 & 0.04 & 0.66 & 0.64 & 0.06 & 0.97 & 0.28 & 0.10 & 0.73 & 0.70 & 0.02 & 0.77 & 0.23 & 0.15 \\
\textbf{IAR} & 0.00 & 0.03 & 0.01 & 0.06 & 0.03 & 0.00 & 0.30 & 0.10 & 0.08 & 0.00 & 0.00 & 0.03 & 0.00 & 0.02 & 0.02 & 0.00 & 0.10 & 0.04 \\
\midrule 
\multicolumn{19}{c}{\textbf{Inter-Context Conflict}} \\
\midrule 
\textbf{OAR} & 0.03 & 0.35 & 0.97 & 0.02 & 0.74 & 0.93 & 0.01 & 0.22 & 0.87 & 0.04 & 0.80 & 0.86 & 0.24 & 0.13 & 0.83 & 0.22 & 0.66 & 0.74 \\
\textbf{CAR} & 0.94 & 0.62 & 0.02 & 0.97 & 0.24 & 0.07 & 0.94 & 0.77 & 0.11 & 0.94 & 0.20 & 0.12 & 0.76 & 0.85 & 0.10 & 0.78 & 0.29 & 0.24 \\
\textbf{IAR} & 0.00 & 0.02 & 0.01 & 0.00 & 0.02 & 0.00 & 0.05 & 0.02 & 0.02 & 0.01 & 0.00 & 0.01 & 0.00 & 0.02 & 0.08 & 0.00 & 0.05 & 0.01 \\
\bottomrule 
\end{tabular}
}
\end{table}

\subsection{Model performance results for different types of knowledge}
% other model + 细分 + 检测
In this section, we show the performance of entity knowledge segmentation. We constructed two entities and three dimensions of knowledge. As shown in Table~\ref{EKopen}, we show the performance of Qwen2.5-VL, InternVL3, and GPT4o-mini models in the face of segmentation knowledge conflicts under open questions.
As shown in Table~\ref{EKMOQ}, we show the performance of Qwen2.5-VL, InternVL3, and GPT4o mini models in the face of segmentation knowledge conflicts under selective questions. It is seen that different types of data show different characteristics. For example, all models are very sensitive to time-related knowledge conflicts. When time conflicts occur, the models tend to be consistent with external documents. While for content-related knowledge conflicts, the models tend to rely on their own memory.Note that we use abbreviations here. PT stands for character time knowledge, PL stands for character nationality knowledge, and PC stands for character occupation knowledge. LT stands for brand time knowledge, LC stands for brand creator knowledge, and LO stands for brand product knowledge.

\section{Detailed examples of three types of multimodal knowledge conflicts in MLLMKC-Bench}
\begin{figure}[b]
    \centering
    \includegraphics[width=1\linewidth]{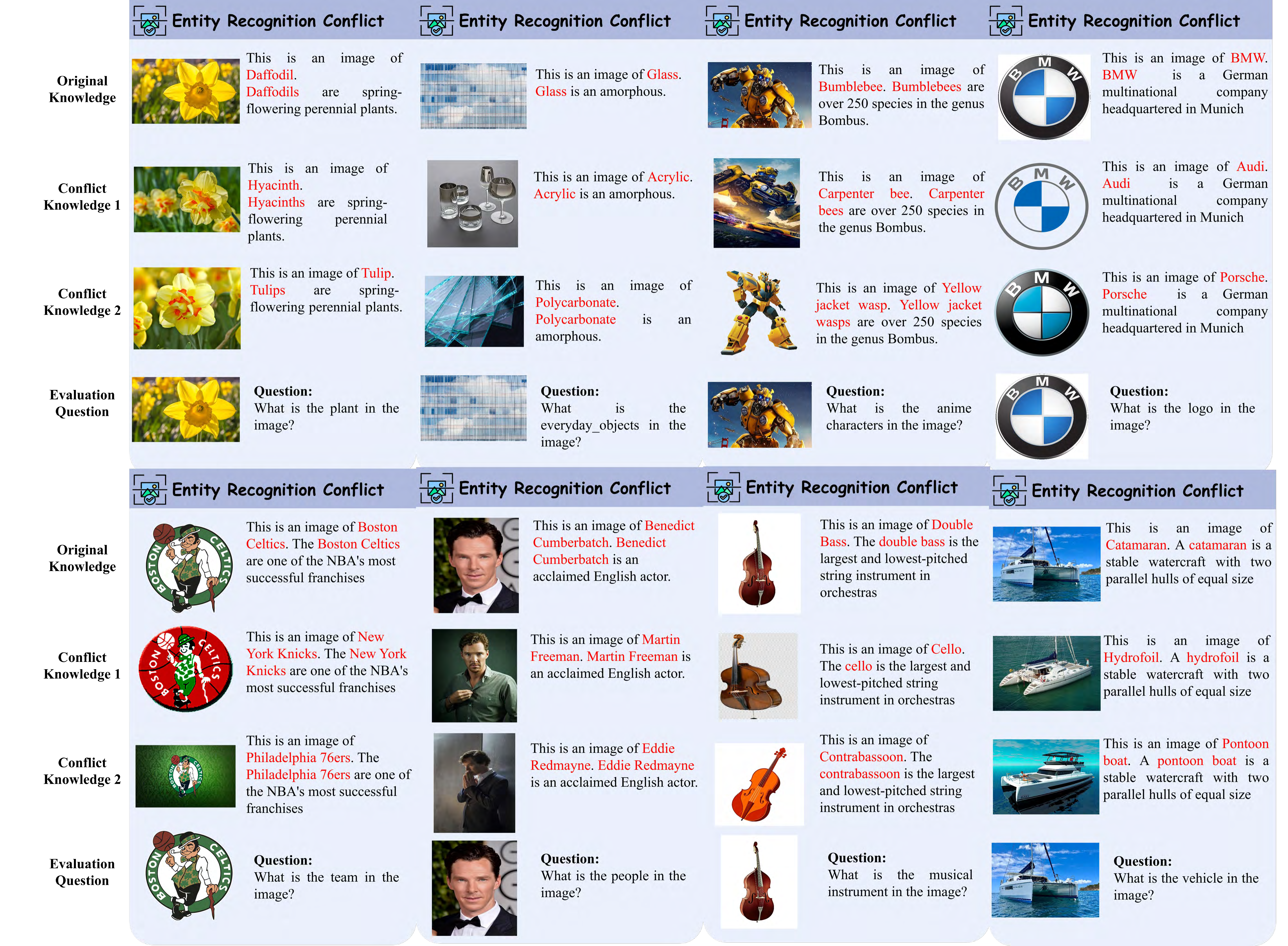}
    \caption{Data instance display diagram of eight entity recognition types.}
    \label{ERR}
\end{figure}

\begin{figure}
    \centering
    \includegraphics[width=1\linewidth]{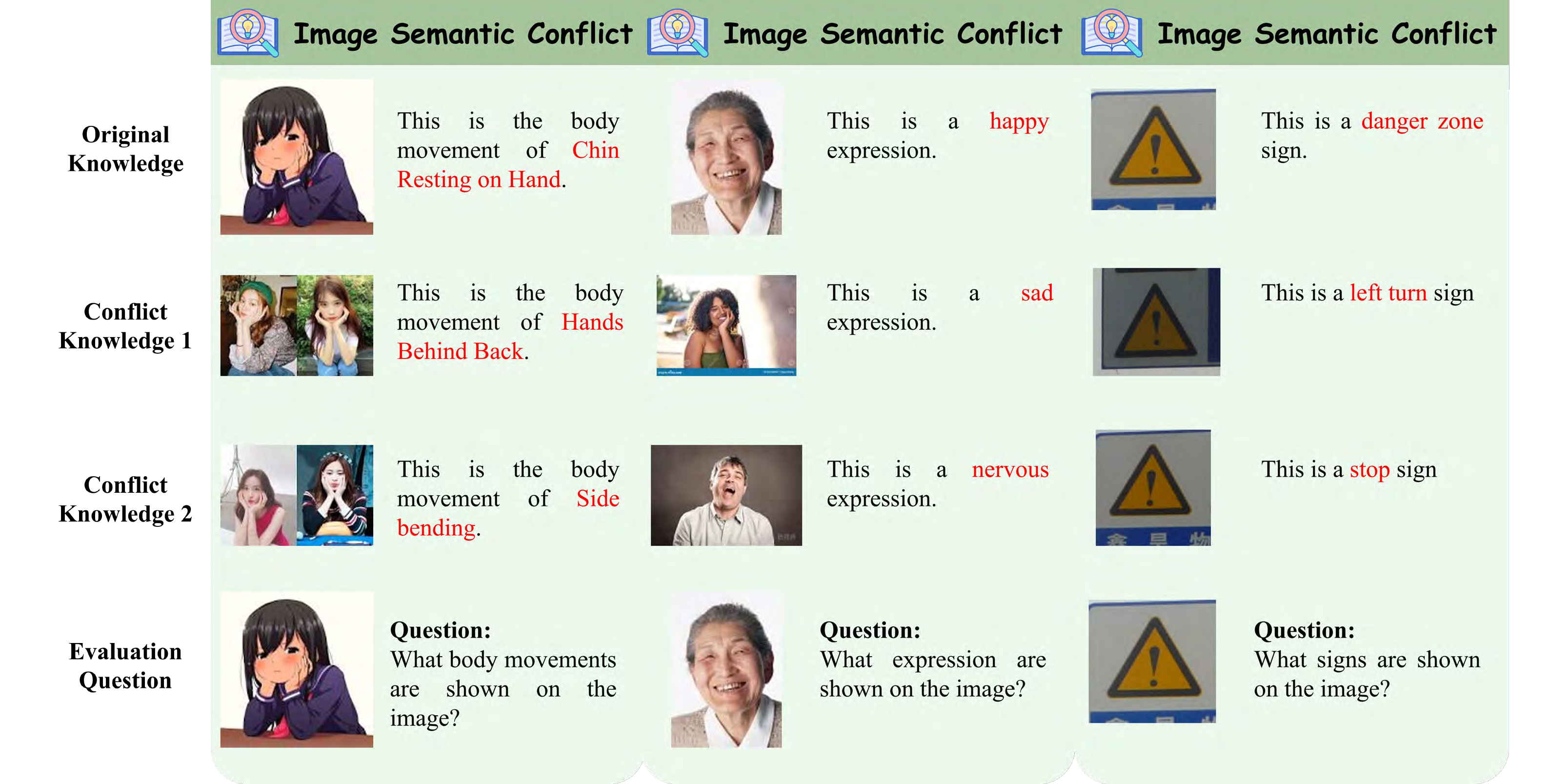}
    \caption{Data instance display diagram of three visual semantic types.}
    \label{VSS}
\end{figure}

\begin{figure}
    \centering
    \includegraphics[width=1\linewidth]{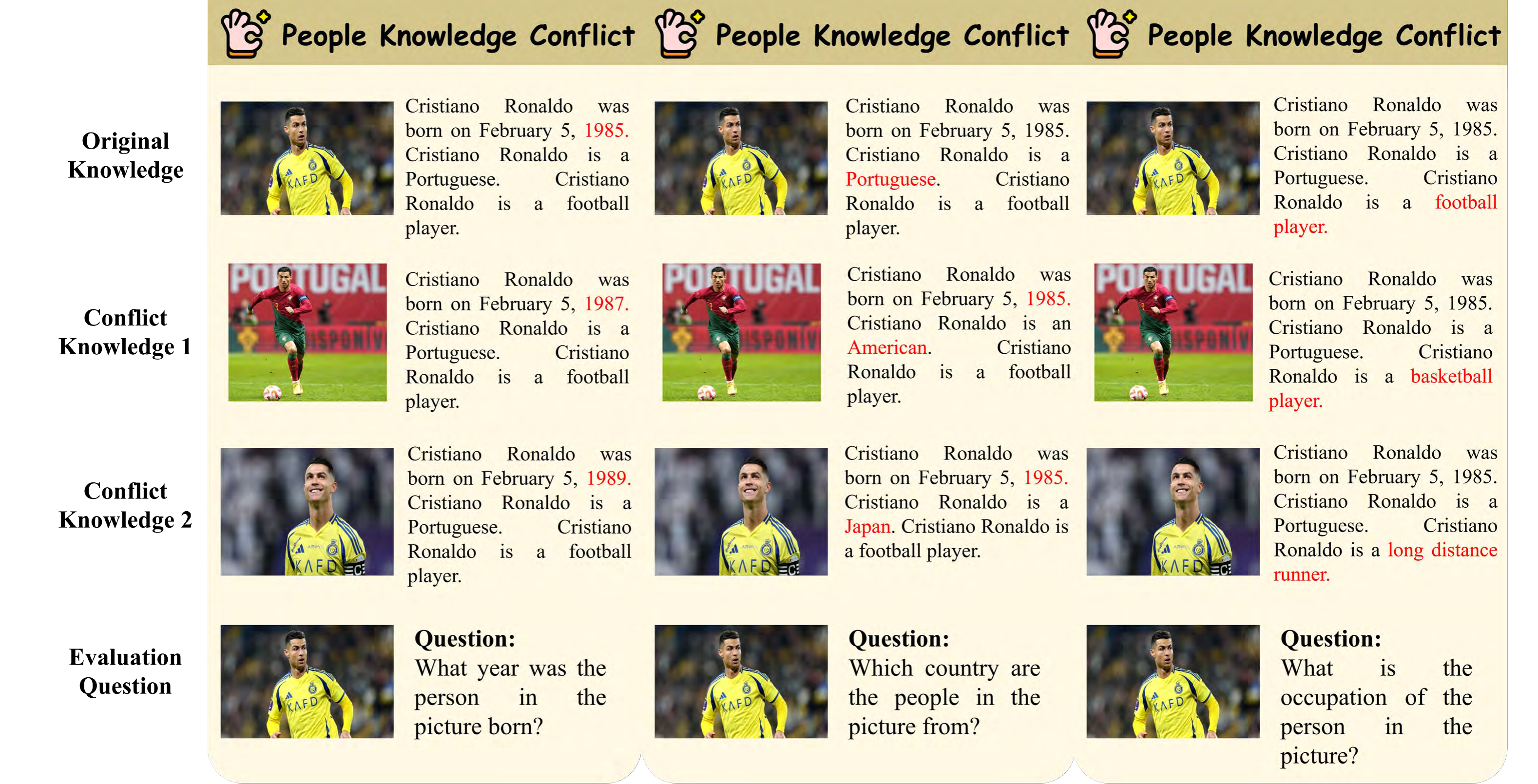}
    \caption{A data example showing three dimensions of character knowledge.}
    \label{PPK}
\end{figure}

\begin{figure}
    \centering
    \includegraphics[width=1\linewidth]{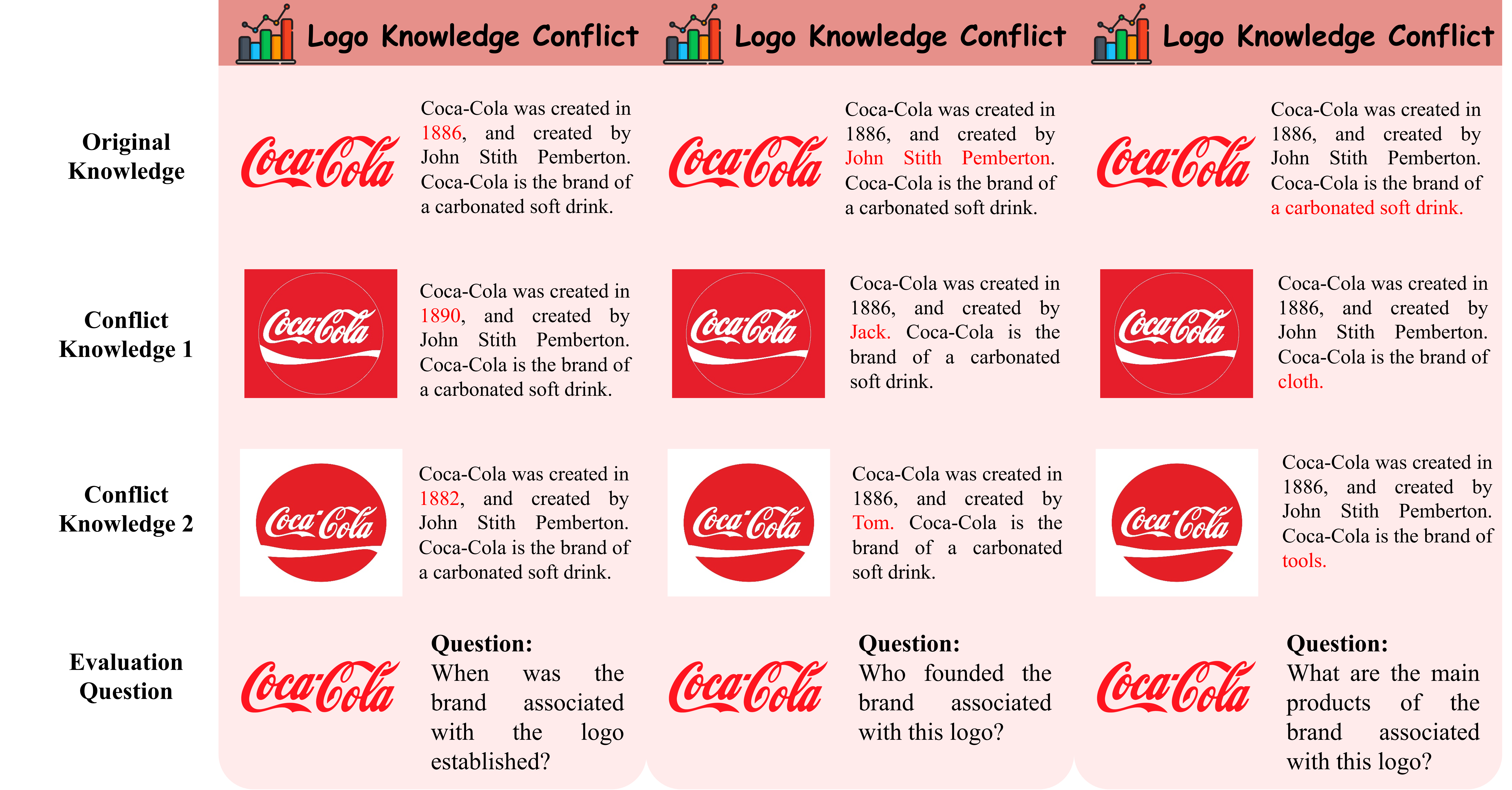}
    \caption{Data example display of three dimensions of brand knowledge.}
    \label{LLK}
\end{figure}

In this section, we present more examples of dataset in detail. For entity recognition, as shown in Figure~\ref{ERR}, we show eight types of data instances: plants, daily necessities, cartoon characters, brand logos, characters, team logos, musical instruments, and transportation. For character knowledge, as shown in Figure \ref{PPK}, we show data instances in three dimensions: character birth time, character nationality, and character occupation. For brand knowledge, as shown in Figure \ref{LLK}, we show data instances in three dimensions: brand creation time, brand creator, and brand main products. For visual semantic, as shown in Figure\ref{VSS},we show three types of data instances: daily actions, expressions, and traffic signs.

\section{Presentation of case studies of different conflict types.}
\begin{figure}
    \centering
    \includegraphics[width=1\linewidth]{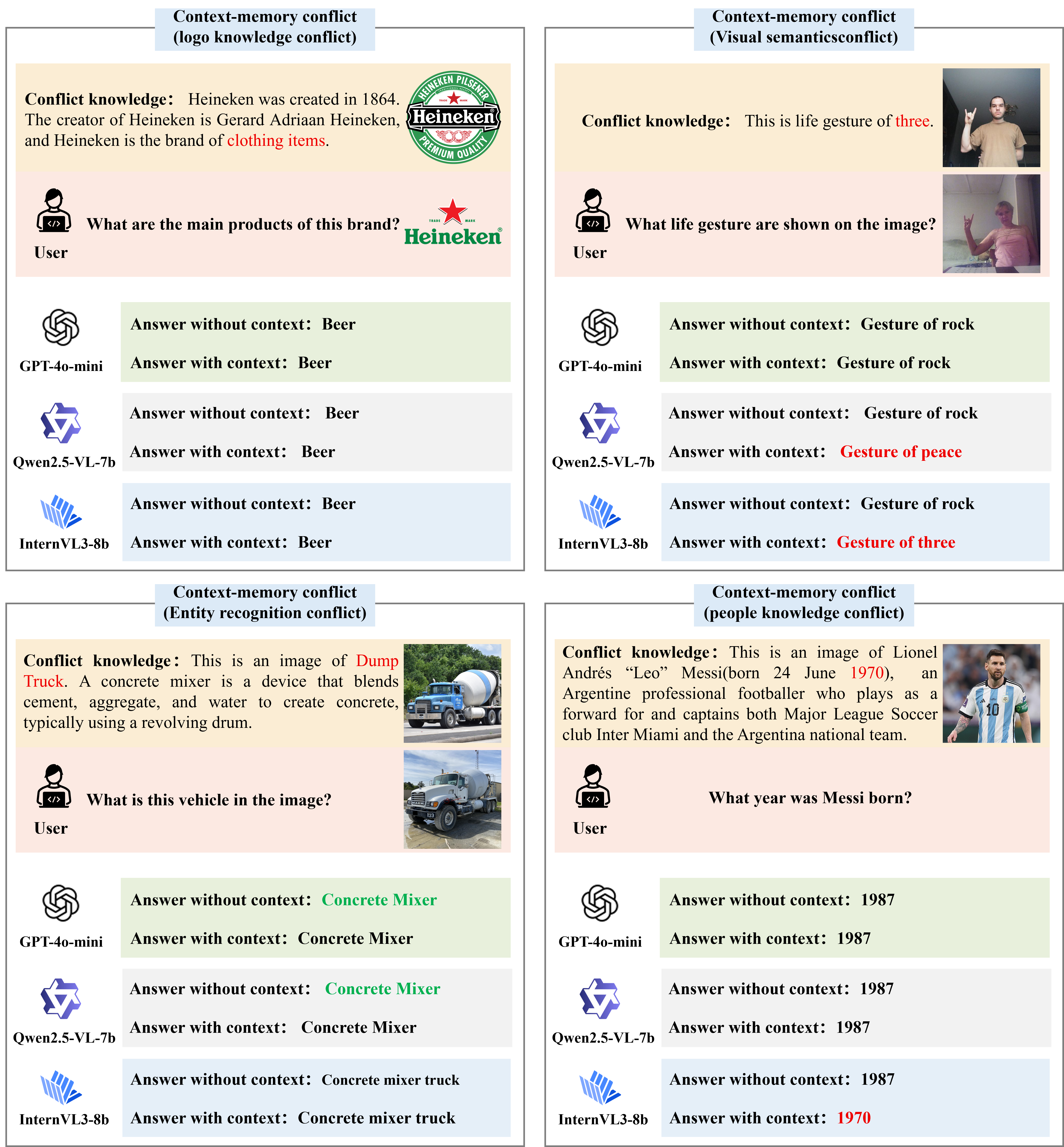}
    \caption{A diagram showing cases of different data instances under Context-Memory Conflict.}
    \label{case11}
\end{figure}

\begin{figure}
    \centering
    \includegraphics[width=1\linewidth]{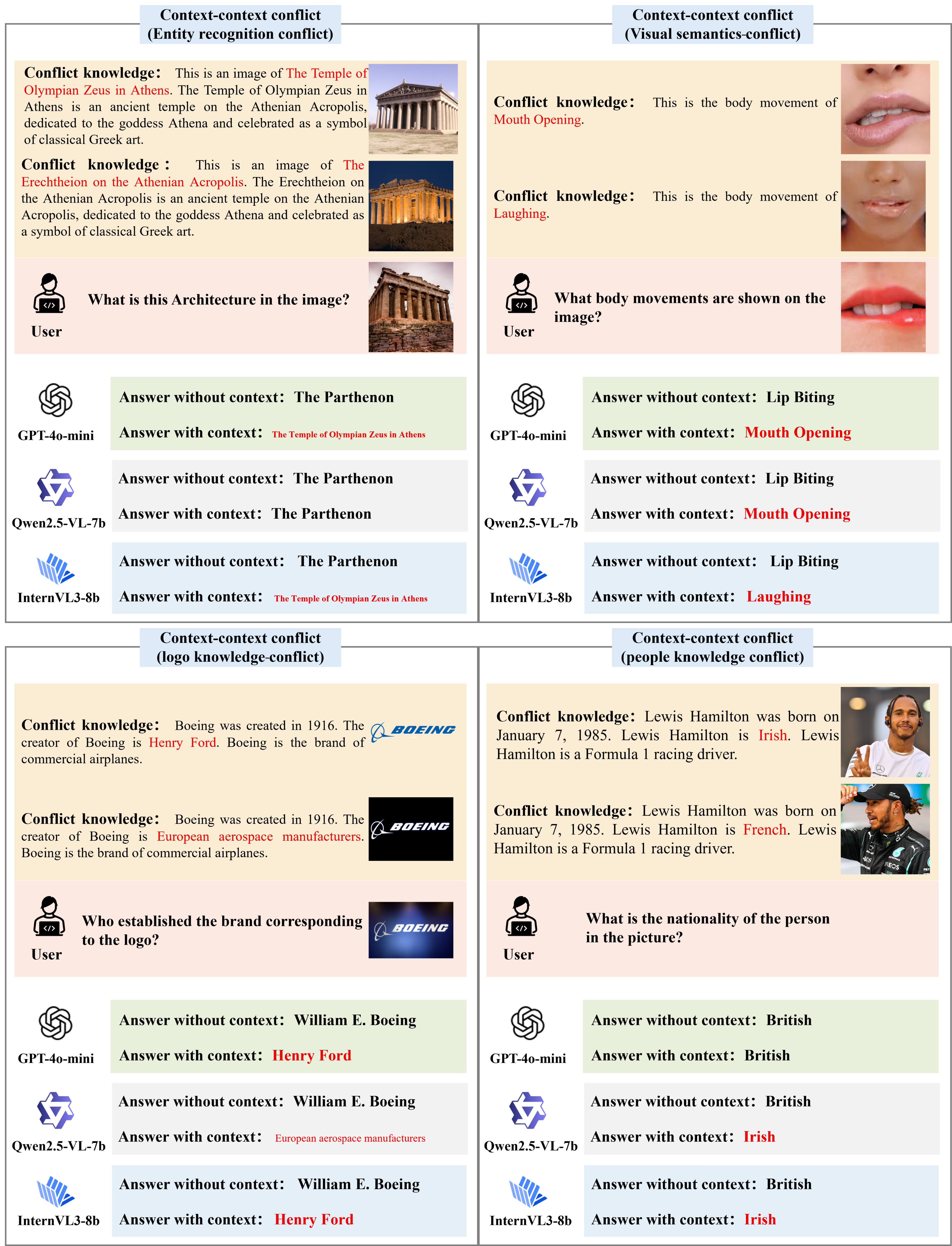}
    \caption{Case diagram showing different data instances under Inter-Context Conflict.}
    \label{case22}
\end{figure}

In this section, we present the case studies of different conflict types. As shown in Figure\ref{case11}, we present the case studies of different data instances when there is Context-Memory Conflict. As shown in Figure\ref{case22}, we present the case studies of different data instances when there is Inter-Context Conflict.

% 2 sample per class

\newpage
\appendix

\end{document}